\documentclass[10pt,journal,compsoc]{IEEEtran}
\usepackage{graphicx}
\usepackage{multirow}
\usepackage{booktabs}
\usepackage{amsmath,amssymb,amsfonts}
\usepackage{algorithm}
\usepackage{algpseudocode}
\usepackage{color}
\usepackage{hyperref}
\ifCLASSOPTIONcompsoc
  \usepackage[nocompress]{cite}
\else
  \usepackage{cite}
\fi

\ifCLASSINFOpdf
\else
\fi

\begin{document}
\title{Contrast-Phys+: Unsupervised and Weakly-supervised Video-based Remote Physiological Measurement via Spatiotemporal Contrast}

\author{Zhaodong~Sun~%\IEEEmembership{Member,~IEEE,}
        and~Xiaobai~Li%,~\IEEEmembership{Life~Fellow,~IEEE}% <-this % stops a space
\IEEEcompsocitemizethanks{\IEEEcompsocthanksitem Z. Sun is with the Center for Machine Vision and Signal Analysis, University of Oulu, Oulu, Finland.
E-mail: zhaodong.sun@oulu.fi
\IEEEcompsocthanksitem X. Li is with The State Key Laboratory of Blockchain and Data Security, Zhejiang University, Hangzhou China, and the Center for Machine Vision and Signal Analysis, University of Oulu, Oulu, Finland. E-mail: xiaobai.li@oulu.fi (Corresponding author: Xiaobai Li)} % <-this % stops an unwanted space
}


\markboth{Journal of \LaTeX\ Class Files,~Vol.~14, No.~8, August~2015}%
{Shell \MakeLowercase{\textit{et al.}}: Bare Demo of IEEEtran.cls for Computer Society Journals}



\IEEEtitleabstractindextext{%
\begin{abstract}
Video-based remote physiological measurement utilizes facial videos to measure the blood volume change signal, which is also called remote photoplethysmography (rPPG). Supervised methods for rPPG measurements have been shown to achieve good performance. However, the drawback of these methods is that they require facial videos with ground truth (GT) physiological signals, which are often costly and difficult to obtain. In this paper, we propose Contrast-Phys+, a method that can be trained in both unsupervised and weakly-supervised settings. We employ a 3DCNN model to generate multiple spatiotemporal rPPG signals and incorporate prior knowledge of rPPG into a contrastive loss function. We further incorporate the GT signals into contrastive learning to adapt to partial or misaligned labels. The contrastive loss encourages rPPG/GT signals from the same video to be grouped together, while pushing those from different videos apart. We evaluate our methods on five publicly available datasets that include both RGB and Near-infrared videos. Contrast-Phys+ outperforms the state-of-the-art supervised methods, even when using partially available or misaligned GT signals, or no labels at all. Additionally, we highlight the advantages of our methods in terms of computational efficiency, noise robustness, and generalization. \textcolor{black}{Our code is available at \url{https://github.com/zhaodongsun/contrast-phys}.}
\end{abstract}

\begin{IEEEkeywords}
Remote Photoplethysmography, Face Video, Unsupervised Learning, Weakly-supervised Learning, Semi-supervised Learning, Contrastive Learning
\end{IEEEkeywords}}

\maketitle

\IEEEdisplaynontitleabstractindextext

% \fi
\IEEEpeerreviewmaketitle

\IEEEraisesectionheading{\section{Introduction}\label{sec:introduction}}


\IEEEPARstart{I}{n} the realm of traditional physiological measurement, skin-contact sensors are commonly employed to capture physiological signals. Examples of such sensors include contact photoplethysmography (PPG) and electrocardiography (ECG). These sensors enable the derivation of crucial physiological parameters such as heart rate (HR), respiration frequency (RF), and heart rate variability (HRV). However, the reliance on skin-contact sensors necessitates specialized biomedical equipment like pulse oximeters, which can lead to discomfort and skin irritation. An alternative approach is remote physiological measurement, which employs cameras to record facial videos for the measurement of remote photoplethysmography (rPPG). This technique harnesses the ability of cameras to capture subtle color changes in the human face, from which multiple physiological parameters including HR, RF, and HRV can be extracted \cite{poh2010advancements}. Unlike traditional methods, video-based physiological measurement relies on readily available cameras rather than specialized biomedical sensors. This approach offers the advantage of not being constrained by physical proximity, rendering it particularly promising for applications in remote healthcare \cite{shi2019atrial,yan2018contact,sun2022non}, emotion analysis \cite{yu2021facial,mcduff2014remote,sabour2021ubfc,sun2022estimating}, and face security \cite{juefei2022countering,ciftci2020fakecatcher,sun2022privacy}.

In earlier studies related to rPPG \cite{verkruysse2008remote,poh2010advancements,de2013robust,wang2016algorithmic}, researchers devised handcrafted features to extract rPPG signals. Subsequently, several deep learning (DL)-based methods \cite{chen2018deepphys,vspetlik2018visual,yu2019remoteBMVC,yu2019remote,lee2020meta,NEURIPS2020_e1228be4,niu2019rhythmnet,niu2020video,lu2021dual,nowara2021benefit} were introduced. These DL-based approaches utilize supervised techniques and diverse network architectures to measure rPPG signals. Under certain conditions, such as when head movements are present or the videos exhibit heterogeneity, DL-based methods tend to exhibit greater robustness compared to traditional handcrafted approaches. However, it's important to note that DL-based rPPG methods heavily rely on extensive datasets comprising face videos and ground truth (GT) physiological signals. Acquiring GT physiological signals, typically measured by contact sensors and synchronized with facial videos, can be a costly endeavor. Issues like missing GT signals or misalignment with facial videos during data collection are common challenges encountered in this context.

Considering the cost and challenges associated with obtaining GT physiological signals, we propose an unsupervised and weakly-supervised method for rPPG measurement, particularly when dealing with data that lacks complete or high-quality labels. The unsupervised method can effectively process facial videos that lack GT signals, while the weakly-supervised method can be employed when dealing with data containing incomplete or low-quality labels, where GT signals may be missing or misaligned.

In our approach, we leverage four key rPPG observations as foundational knowledge. 1) \textbf{rPPG spatial similarity}: rPPG signals obtained from different facial regions tend to exhibit similar power spectrum densities (PSDs). 2) \textbf{rPPG temporal similarity}: Segments of rPPG data taken within short time intervals (e.g., two consecutive 5-second clips) typically display similar PSDs, as HR tends to transition smoothly in most cases. 3) \textbf{Cross-video rPPG dissimilarity}: PSDs of rPPG signals from different videos often exhibit variations. 4) \textbf{HR range constraint}: The HR typically falls within the range of 40 to 250 beats per minute (bpm).

In our prior ECCV 2022 publication \cite{sun2022contrast}, we introduced Contrast-Phys, an unsupervised learning framework. The present work, referred to as Contrast-Phys+, represents an extension of our earlier research. This work contributes significantly in the following ways:

\begin{itemize}

\item We propose Contrast-Phys+, a versatile model capable of adapting to diverse data conditions, including scenarios with no labels, partial labels, or misaligned labels. Importantly, Contrast-Phys+ operates effectively in both unsupervised and weakly-supervised settings. To the best of our knowledge, Contrast-Phys+ is the first work to train an rPPG model in both weakly-supervised and unsupervised settings.

\item We showcase the efficacy of Contrast-Phys+ in weakly-supervised scenarios, where some ground truth signals may be missing or lack synchronization. Remarkably, Contrast-Phys+ with missing labels exhibits performance that can surpass that of fully supervised methods employing complete label sets. Moreover, Contrast-Phys+ demonstrates significantly enhanced robustness when faced with ground truth signal desynchronization, outperforming other fully supervised methods.

\item We conduct extensive experiments and analyses pertaining to Contrast-Phys+. A comprehensive performance comparison is also offered, contrasting the capabilities of Contrast-Phys+ against recent state-of-the-art baselines. Additional experiments also demonstrate that Contrast-Phys+ can use unlabeled data to expand and diversify the training dataset for improved generalization. We also offer a thorough analysis of the reasons why Contrast-Phys+ can be effective in unsupervised and weakly-supervised scenarios. Besides, we present statistical analysis to validate the proposed rPPG observations and include detailed ablation studies to substantiate the effectiveness of Contrast-Phys+. 

\end{itemize}

%\IEEEpubidadjcol

\section{Related Work}

\subsection{Video-Based Remote Physiological Measurement}

The concept of measuring remote photoplethysmography (rPPG) from facial videos via the green channel was initially introduced by Verkruysse et al. \cite{verkruysse2008remote}. Subsequently, various handcrafted methods \cite{poh2010advancements,de2013robust,de2014improved,wang2016algorithmic,lam2015robust,tulyakov2016self,wang2014exploiting,nowara2020near} were proposed to enhance the quality of rPPG signals. These methods, predominantly developed in the earlier years, relied on manual procedures and did not necessitate training datasets, earning them the label of "traditional methods." In recent years, deep learning (DL) techniques have surged in rPPG measurement. Some studies \cite{chen2018deepphys,vspetlik2018visual,NEURIPS2020_e1228be4,nowara2021benefit,li2023learning} employed a 2D convolutional neural network (2DCNN) with two consecutive video frames as input for rPPG estimation. Another category of DL-based methods \cite{niu2019rhythmnet,niu2020video,lu2021dual,lu2023neuron,du2023dual} utilized spatial-temporal signal maps extracted from various facial regions as input for 2DCNN models. Additionally, 3DCNN-based methods \cite{yu2019remote,yu2019remoteBMVC,gideon2021way} were introduced to achieve high performance, particularly on compressed videos \cite{yu2019remote}. These DL-based approaches, categorized as supervised methods, demand both facial videos and ground truth (GT) physiological signals for training. More recently, Wang et al. \cite{wang2022self} proposed a self-supervised rPPG method to acquire rPPG representations, although it still necessitates heart rate (HR) labels for fine-tuning the rPPG model. Gideon et al. \cite{gideon2021way} introduced the first unsupervised rPPG method, which does not rely on GT physiological signals for training. However, this method, while pioneering, exhibits lower accuracy compared to state-of-the-art supervised methods and can be sensitive to external noise. Subsequent to these developments, multiple unsupervised rPPG techniques have emerged \cite{sun2022contrast,speth2023non,yang2022simper,yue2023facial}. These unsupervised rPPG methods have gained attention because they solely require facial videos for training, eliminating the need for GT signals, yet they achieve performance levels similar to those of supervised methods. This is particularly advantageous given the expense associated with collecting GT signals alongside facial videos. However, none of the methods above considered utilizing partial or low-quality labels to further refine rPPG signal quality.

\subsection{Contrastive Learning}

Contrastive learning, a widely adopted self-supervised learning technique in computer vision tasks, empowers deep learning models to map high-dimensional images or videos into lower-dimensional feature embeddings without the need for labeled data \cite{hadsell2006dimensionality,schroff2015facenet,van2018representation,chen2020simple,tian2020contrastive,he2020momentum,grill2020bootstrap,misra2020self,qian2021spatiotemporal}. Its primary objective is to ensure that features derived from different perspectives of the same sample (referred to as positive pairs) are brought closer together, while features from different samples (referred to as negative pairs) are pushed apart. This approach finds extensive utility in pre-training models, thereby facilitating subsequent task-specific training in domains such as image classification \cite{chen2020simple}, video analysis \cite{qian2021spatiotemporal,qi2021self}, face recognition \cite{schroff2015facenet}, and face detection \cite{wang2022unsupervised}. This is particularly advantageous in situations characterized by limited access to labeled data. In our research, we leverage prior knowledge related to remote photoplethysmography (rPPG) to generate suitable positive and negative pairs of rPPG signal instances for contrastive learning. Diverging from prior methodologies that focus on feature embedding, our proposed method, Contrast-Phys+, possesses the capability to directly generate rPPG signals without the need for labeled data, thereby enabling unsupervised learning. Additionally, we harness ground truth (GT) signals to construct positive/negative pairs for contrastive learning, thus facilitating end-to-end weakly-supervised training even in scenarios where labels are missing or of suboptimal quality.

\section{Observations about rPPG}\label{sec:observation}
This section describes four observations about rPPG, which are the preconditions to designing Contrast-Phys+ and enabling unsupervised and weakly-supervised learning.

\subsection{rPPG Spatial Similarity} 

\begin{figure*}[hbt!]
\centering
\begin{minipage}[b]{0.9\linewidth}
  \centering
  \centerline{\includegraphics[width=\linewidth]{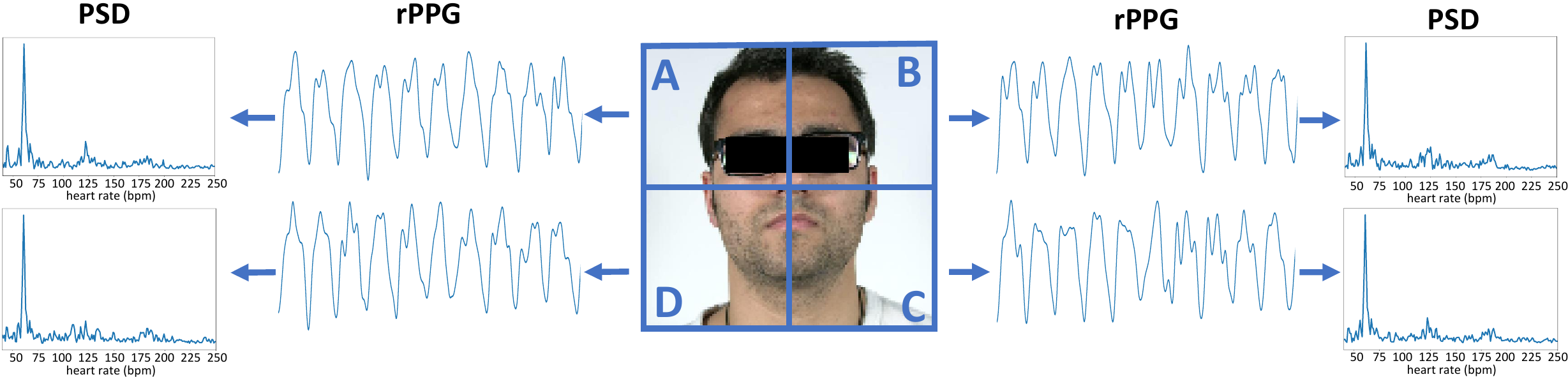}}
\end{minipage}
\caption{Illustration of rPPG spatial similarity. The rPPG signals from four facial areas (A, B, C, D) have similar waveforms and power spectrum densities (PSDs).}
\label{fig:rppg_spatial}
\end{figure*}

rPPG signals originating from various facial regions exhibit analogous waveforms, accompanied by the similarity in their Power Spectrum Densities (PSDs). This spatial coherence in rPPG signals has been leveraged in the design of multiple methodologies, as demonstrated in prior works \cite{lam2015robust,tulyakov2016self,kumar2015distanceppg,wang2014exploiting,wang2019discriminative,liu20163d,liu2018remote}. While subtle phase and amplitude disparities may exist in the temporal domain when comparing rPPG signals from distinct skin areas \cite{kamshilin2011photoplethysmographic,kamshilin2013variability}, these distinctions become inconsequential when rPPG waveforms are analyzed in the frequency domain, where PSDs are normalized. As illustrated in Fig. \ref{fig:rppg_spatial}, the rPPG waveforms derived from four distinct spatial regions share a striking resemblance, characterized by identical peaks in their respective PSDs.

\subsection{rPPG Temporal Similarity} \label{sec:temp_similarity}

\begin{figure}[hbt!]
\centering
\begin{minipage}[b]{\linewidth}
  \centering
  \centerline{\includegraphics[width=\linewidth]{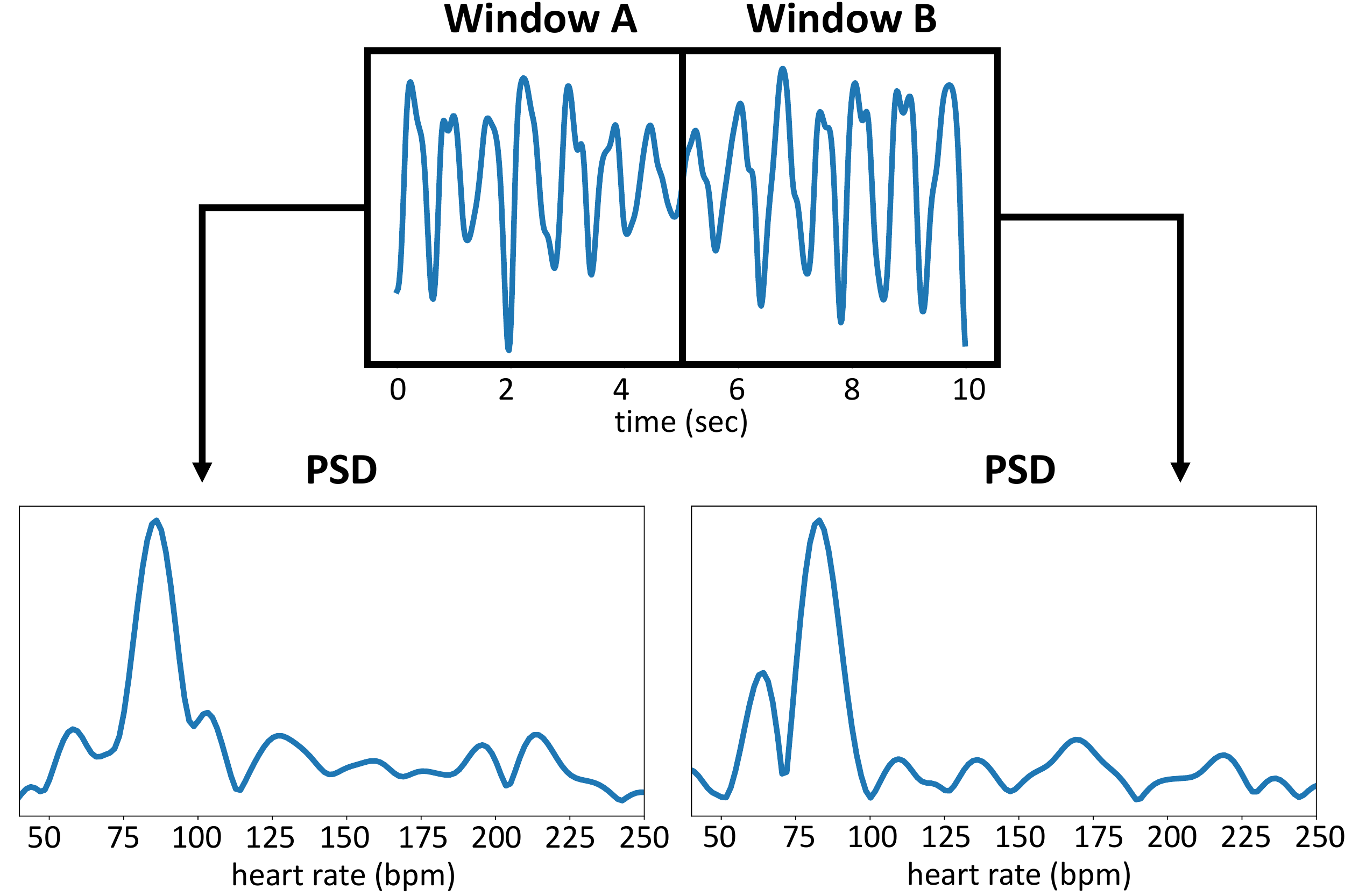}}
\end{minipage}
\caption{Illustration of rPPG temporal similarity. The rPPG signals from two temporal windows (A, B) have similar PSDs.}
\label{fig:rppg_temporal}
\end{figure}

The heart rate (HR) undergoes gradual changes within short time frames, as noted by Gideon et al. \cite{gideon2021way}. A similar finding was reported by Stricker et al. \cite{stricker2014non}, who observed slight HR variations in their dataset over short time intervals. Given that HR is prominently represented by a dominant peak in the PSD, it follows that the PSD experiences minimal fluctuations as well. Therefore, when randomly selecting small temporal windows from a brief rPPG segment (e.g., 10 seconds), one can anticipate that the PSDs of these windows will exhibit similarity. As depicted in Fig. \ref{fig:rppg_temporal}, we illustrate this by sampling two 5-second windows from a 10-second rPPG signal and comparing the PSDs of these windows. Indeed, the two PSDs demonstrate similarity, with dominant peaks occurring at identical frequencies. It is important to note that this observation holds true when dealing with short-term rPPG signals. We will delve into the impact of signal duration on our model's performance in the forthcoming ablation study.

We can summarize spatiotemporal rPPG similarity using the following \textcolor{black}{relation}.

\begin{equation}
\resizebox{0.5\textwidth}{!}{
$\begin{aligned}
\text{PSD}\big\{ G\big(v(t_1 \to t_1 + \Delta t, \mathcal{H}_1, \mathcal{W}_1)\big) \big\} \approx \text{PSD}\big\{ G\big(v(t_2 \to t_2 + \Delta t, \mathcal{H}_2, \mathcal{W}_2)\big) \big\}
\end{aligned}$
}
\end{equation}

In this \textcolor{black}{relation}, $v \in \mathbb{R}^{T \times H \times W \times 3}$ represents a facial video, and $G$ signifies an rPPG measurement algorithm. We can select a facial region defined by height $\mathcal{H}_1$ and width $\mathcal{W}_1$ and a time interval $t_1 \to t_1+\Delta t$ from video $v$ to derive one rPPG signal. A similar rPPG signal can be obtained from the same video, utilizing parameters $\mathcal{H}_2$, $\mathcal{W}_2$, and $t_2 \to t_2+\Delta t$. To meet the criteria for short-term rPPG signals, the temporal separation $\vert t_1 - t_2 \vert$ should remain small.

\subsection{Cross-video rPPG Dissimilarity}

\begin{figure}[hbt!]
\centering
\begin{minipage}[b]{\linewidth}
  \centering
  \centerline{\includegraphics[width=\linewidth]{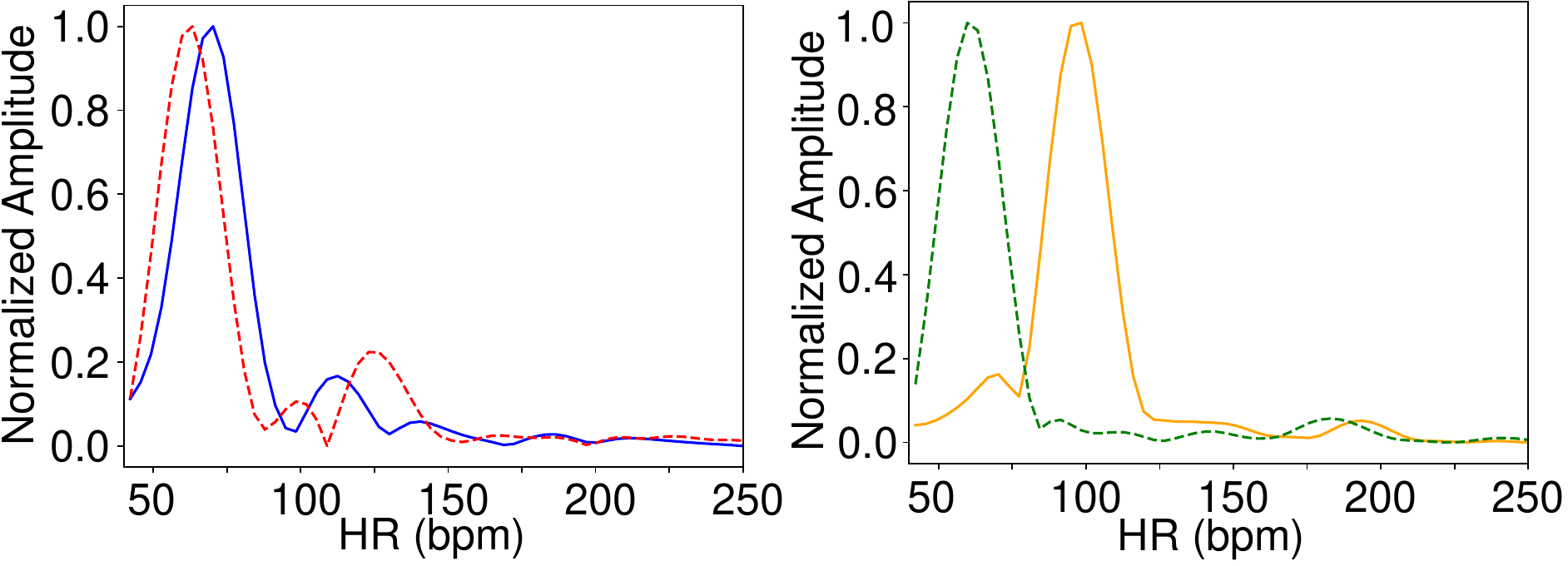}}
\end{minipage}
\caption{The most similar (left) and most different (right) cross-video PSD pairs in the OBF dataset.}
\label{fig:cross-video}
\end{figure}

rPPG signals obtained from different facial videos exhibit distinct PSDs. This divergence arises from the fact that each video features distinct individuals with varying physiological conditions, such as physical activity and emotional states, which are known to influence HRs \cite{hr_factor}. Even in cases where HRs between two videos may appear similar, disparities in the PSDs can persist. This is due to the presence of additional physiological factors within the PSDs, such as respiration rate \cite{chen2020modulation} and HRV \cite{pai2021hrvcam}, which are unlikely to align entirely across different videos. To substantiate this observation, we conducted an analysis involving the calculation of mean squared errors for cross-video PSD pairs within the OBF dataset \cite{li2018obf}. The results, illustrated in Fig. \ref{fig:cross-video}, underscore the primary dissimilarity in cross-video PSDs as being centered around the heart rate peak.

This cross-video rPPG dissimilarity is described by the following \textcolor{black}{relation}: 

\begin{equation}
\resizebox{0.5\textwidth}{!}{
$\begin{aligned}
\text{PSD}\big\{ G\big(v(t_1 \to t_1 + \Delta t, \mathcal{H}_1, \mathcal{W}_1)\big) \big\} \neq \text{PSD}\big\{ G\big(v^{\prime}(t_2 \to t_2 + \Delta t, \mathcal{H}_2, \mathcal{W}_2)\big) \big\}
\end{aligned}$
}
\end{equation}

where $v$ and $v^{\prime}$ represent two distinct videos. By selecting specific facial areas and time intervals from these videos, one can expect the PSDs of the two resulting rPPG signals to exhibit noticeable differences.

\subsection{HR Range Constraint} \label{sec:hr_range_constraint}
The typical HR range for the majority of individuals falls within the interval of 40 to 250 beats per minute (bpm) \cite{hr_interval}. In line with established practices \cite{poh2010advancements,li2014remote}, this HR range serves as the basis for rPPG signal filtering, with the highest peak identified within this range to estimate HR. Consequently, our method will primarily concentrate on PSD within the frequency band of 0.66 Hz to 4.16 Hz.

\section{Method}

\begin{figure*}[hbt!]
\centering
\begin{minipage}[b]{0.95\linewidth}
  \centering
  \centerline{\includegraphics[width=\linewidth]{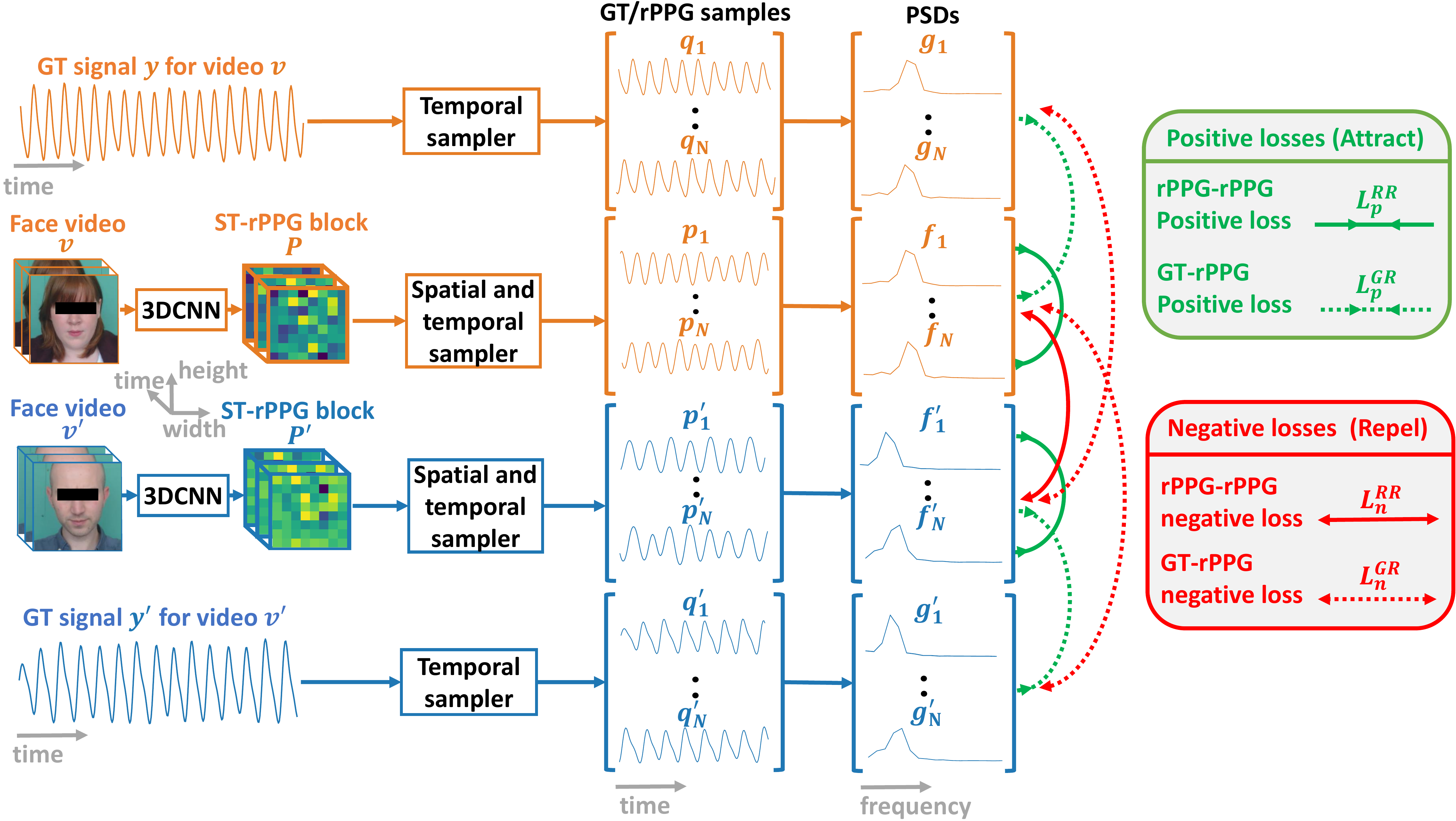}}
\end{minipage}
\caption{The diagram of Contrast-Phys+ for weakly-supervised or unsupervised learning.}
\label{fig:diagram_new}
\end{figure*}

In this section, we propose Contrast-Phys+ for weakly-supervised and unsupervised rPPG learning as shown in Fig. \ref{fig:diagram_new}. We describe the face preprocessing in Sec. \ref{sec:preprocessing}, the ST-rPPG block representation in Sec. \ref{sec:ST-rPPG block representation}, the rPPG spatiotemporal sampling in Sec. \ref{sec:rppg sampling}, and the contrastive loss function in Sec. \ref{sec:contrastive loss function}. 

\subsection{Preprocessing}\label{sec:preprocessing}
The initial step involves preprocessing the original video, and the primary task is facial cropping. Utilizing OpenFace \cite{baltrusaitis2018openface}, we generate facial landmarks. To determine the central facial point for each frame, we compute the minimum and maximum horizontal and vertical coordinates of these landmarks. Subsequently, a bounding box is established, sized at 1.2 times the vertical coordinate range of the landmarks observed in the initial frame, and this size remains constant for all subsequent frames. With the central facial point and bounding box size determined for each frame, we proceed to crop the face in every frame. These cropped facial regions are then resized to dimensions of $128 \times 128$, rendering them ready for input into our model.

\subsection{Spatiotemporal rPPG (ST-rPPG) Block Representation}\label{sec:ST-rPPG block representation}

We have adapted the 3DCNN-based PhysNet \cite{yu2019remoteBMVC} to compute the ST-rPPG block representation. Our modified model takes as input an RGB video with dimensions $T \times 128 \times 128 \times 3$, where $T$ represents the number of frames. In the final stage of our model, we employ adaptive average pooling to perform downsampling along spatial dimensions, enabling control over the output spatial size. This alteration facilitates the generation of a spatiotemporal rPPG block with dimensions $T \times S \times S$, where $S$ denotes the length of the spatial dimension, as depicted in Fig. \ref{fig:sampler}. Further elaboration on the 3DCNN model is available in the supplementary material.

The ST-rPPG block is essentially a collection of rPPG signals embedded within spatiotemporal dimensions. To denote this ST-rPPG block, we use $P \in \mathbb{R}^{T \times S \times S}$. When selecting a specific spatial location $(h, w)$ within the ST-rPPG block, the corresponding rPPG signal $P(\cdot, h, w)$ is extracted from the receptive field associated with that spatial position in the input video. It is worth noting that when the spatial dimension length $S$ is small, each spatial position within the ST-rPPG block encompasses a larger receptive field, albeit with fewer rPPG signals contained within the block. Importantly, the receptive field of each spatial position in the ST-rPPG block encompasses a portion of the facial region, ensuring that all spatial positions in the ST-rPPG block encompass valuable rPPG information.

\subsection{rPPG Spatiotemporal Sampling}\label{sec:rppg sampling}

\begin{figure}[hbt!]
\centering
\begin{minipage}[b]{\linewidth}
  \centering
  \centerline{\includegraphics[width=\linewidth]{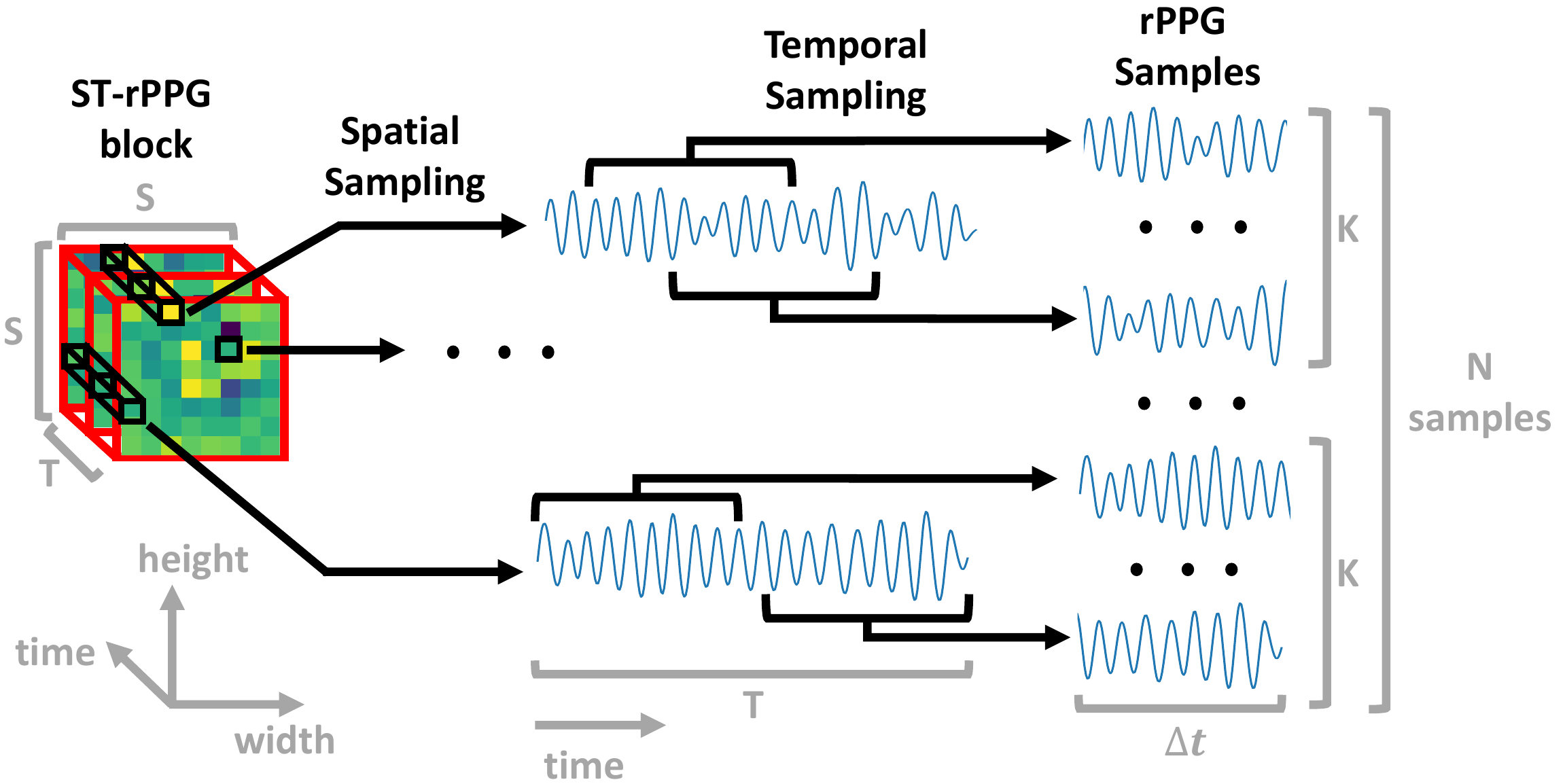}}
\end{minipage}
\caption{Spatial and temporal Sampler for an ST-rPPG Block.}
\label{fig:sampler}
\end{figure}

\begin{algorithm}
  \label{alg:sampling}
  \textbf{Input}: ST-rPPG block: $P$ with shape $T \times S \times S$, Number of rPPG samples per spatial location: $K$, The default rPPG sample length $\Delta t = T/2$
\begin{algorithmic}[1]
\caption{rPPG Spatiotemporal Sampling}
\State Initialze an empty list $H$ for storing all rPPG samples

\For{$h, w \in \{1,...,S\}, \{1,...,S\}$} \Comment Loop over all spatial locations
\For{$k \in \{1,...,K\}$} \Comment $K$ rPPG samples per spatial location
\State Randomly choose a starting time $t$ between 0 and $T-\Delta t$
\State Append the rPPG sample $P(t \to t+\Delta t, h, w)$ into the list $H$
\EndFor
\EndFor
\end{algorithmic}
\textbf{Output}: The list $H=[p_1,...,p_N]$ containing rPPG samples
\end{algorithm}

In the process of generating rPPG samples from the ST-rPPG block, as depicted in Fig. \ref{fig:sampler}, which is the spatial and temporal sampler illustrated in Fig. \ref{fig:diagram_new}, we employ both spatial and temporal sampling techniques. For spatial sampling, we extract the rPPG signal denoted as $P(\cdot, h, w)$ from a specific spatial position. In the case of temporal sampling, we select a short time interval from $P(\cdot, h, w)$, resulting in the final spatiotemporal sample, denoted as $P(t \to t+\Delta t, h, w)$, where $h$ and $w$ represent the spatial position, $t$ signifies the starting time, and $\Delta t$ signifies the duration of the time interval.

Given an ST-rPPG block, we iterate through all spatial positions and extract $K$ rPPG clips, each with a randomly selected starting time $t$, for each spatial position as shown in Algorithm 1. Consequently, we obtain a total of $N = S \cdot S \cdot K$ rPPG clips from the ST-rPPG block. It is important to note that these sampling procedures are employed to generate multiple rPPG samples for use in contrastive learning during the model training phase. During inference, the ST-rPPG block is spatially averaged to yield the final rPPG signal.

\subsection{GT Signal Temporal Sampling} \label{sec:gt sampling}

\begin{figure}[hbt!]
\centering
\begin{minipage}[b]{0.9\linewidth}
  \centering
  \centerline{\includegraphics[width=\linewidth]{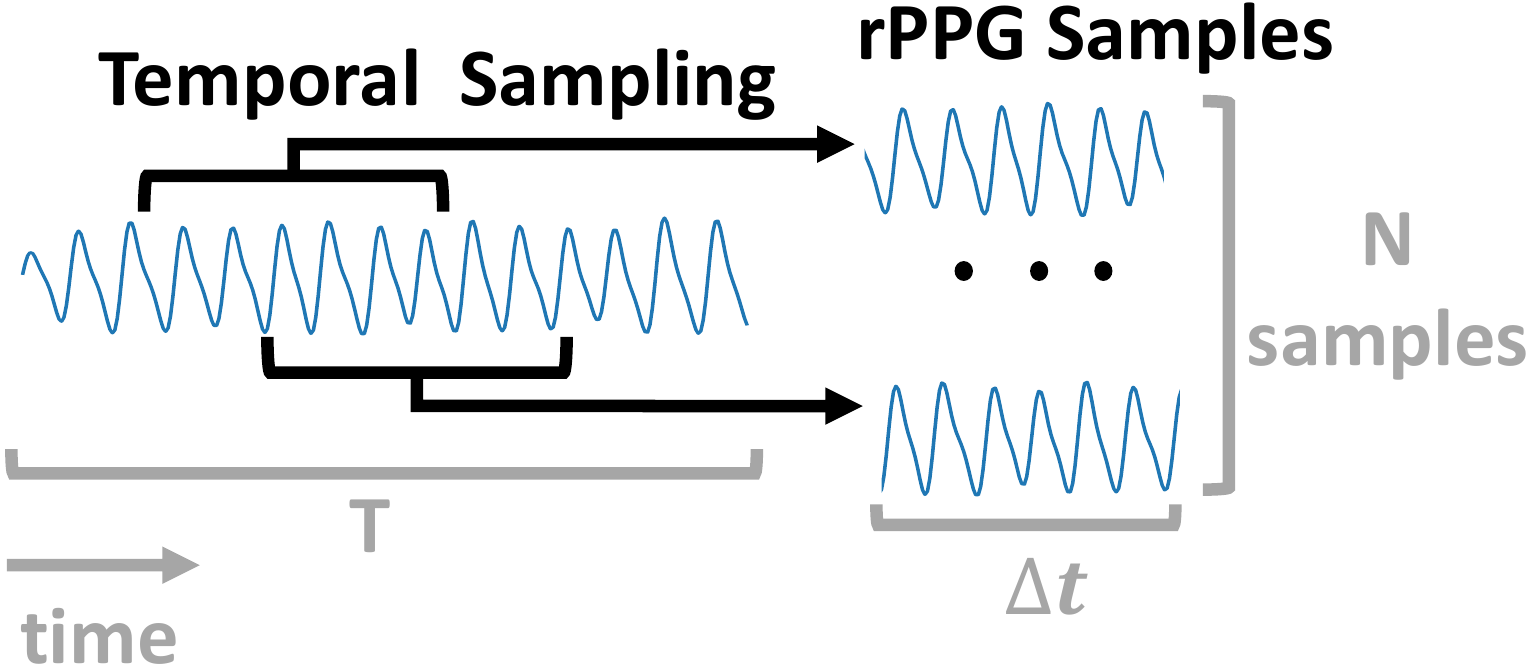}}
\end{minipage}
\caption{Temporal Sampler for a GT Signal.}
\label{fig:sampler_T}
\end{figure}
 
Unlike ST-rPPG blocks, which encompass spatiotemporal signals, GT signals, which are one-dimensional temporal signals, necessitate different sampling approaches. Given the dimensional disparity between GT signals and ST-rPPG blocks, we employ temporal sampling for GT signals and spatiotemporal sampling for ST-rPPG blocks. The GT signal temporal sampling process, as depicted in Fig. \ref{fig:sampler_T}, entails selecting a short time interval from the GT signal $y$, resulting in a temporal sample denoted as $y(t \to t+\Delta t)$, where $t$ represents the starting time, and $\Delta t$ signifies the duration of the time interval. For a single GT signal, we sample $N$ GT clips, each with a randomly determined starting time $t$.

As illustrated in both the top and bottom branches of Fig. \ref{fig:diagram_new}, the GT signals $y$ and $y'$ corresponding to the facial videos $v$ and $v'$ undergo temporal sampling, generating two sets of GT samples, namely $[q_1,...,q_N]$ and $[q'_1,...,q'_N]$, respectively. Subsequently, these two sets of GT samples are transformed into two sets of PSDs, denoted as $[g_1,...,g_N]$ and $[g'_1,...g'_N]$, respectively.

\subsection{Contrastive Loss for Contrast-Phys+}\label{sec:contrastive loss function}

As illustrated in Fig. \ref{fig:diagram_new}, our process begins with the selection of two distinct videos randomly chosen from a dataset as input. For each video, we derive an ST-rPPG block denoted as $P$, a set of rPPG samples $[p_1, \dots, p_N]$, and their corresponding rPPG PSDs $[f_1, \dots, f_N]$. If the GT signal is available for this video, we additionally obtain a set of GT signal samples $[q_1, \dots, q_N]$ and their corresponding GT PSDs $[g_1, \dots, g_N]$. This procedure is repeated for the second video. Note that we normalize each PSD by dividing it by its summation, ensuring all PSDs are on the same scale.

As shown in Fig. \ref{fig:diagram_new} (right), the underlying principle of our contrastive loss lies in the alignment of GT-rPPG or rPPG-rPPG PSD pairs stemming from the same video, while simultaneously pushing apart GT-rPPG or rPPG-rPPG PSD pairs originating from different videos. The HR range constraint is imposed after the rPPG signals are converted to PSDs. We only keep PSD values within the HR range of 0.66 Hz to 4.16 Hz (corresponding to 40 to 250 beats per minute) while removing the PSD values outside the HR range, so only the PSD values within the HR range are used in our method. The HR range constraint has no impact on the input videos.

\textbf{rPPG-rPPG Positive Loss.} In accordance with the rPPG spatiotemporal similarity, it is expected that the rPPG PSDs resulting from spatiotemporal sampling of the same ST-rPPG block should exhibit similarity. The following \textcolor{black}{relations} outline this property for the two input videos:

For one video:
\begin{equation}
\resizebox{0.5\textwidth}{!}{
$\begin{aligned}
&\text{PSD}\big\{P(t_1 \to t_1+\Delta t, h_1, w_1)\big\} \approx \text{PSD}\big\{P(t_2 \to t_2+\Delta t, h_2, w_2)\big\} 
\\
&\implies f_i \approx f_j, i \neq j
\end{aligned}$
}
\end{equation}

For the other video:
\begin{equation}
\resizebox{0.5\textwidth}{!}{
$\begin{aligned}
&\text{PSD}\big\{P^{\prime}(t_1 \to t_1+\Delta t, h_1, w_1)\big\} \approx \text{PSD}\big\{P^{\prime}(t_2 \to t_2+\Delta t, h_2, w_2)\big\} 
\\
&\implies f^{\prime}_i \approx f^{\prime}_j, i \neq j
\end{aligned}$
}
\end{equation}

To bring together the rPPG PSDs from the same video, we employ the mean squared error as the loss function for rPPG-rPPG positive pairs, denoted as $(f_i,f_j)$. The rPPG-rPPG positive loss term, $L_p^{RR}$, is presented below, and it is normalized with respect to the total number of rPPG-rPPG positive pairs.

\begin{equation}
L_p^{RR} =  \sum_{i=1}^{N} \sum_{\substack{j=1 \\j \neq i}}^{N} 
\frac{\parallel f_i - f_j \parallel^2 + \parallel f^{\prime}_i - f^{\prime}_j \parallel^2}{2N(N-1)}
\end{equation}

\textbf{rPPG-rPPG Negative Loss.} In accordance with the cross-video rPPG dissimilarity, it is expected that the rPPG PSDs resulting from spatiotemporal sampling of two different ST-rPPG blocks should differ. We can employ the following \textcolor{black}{relation} to describe this property for the two input videos:

\begin{equation}
\resizebox{0.5\textwidth}{!}{
$\begin{aligned}
&\text{PSD}\big\{P(t_1 \to t_1+\Delta t, h_1, w_1)\big\} \neq \text{PSD}\big\{P^{\prime}(t_2 \to t_2+\Delta t, h_2, w_2)\big\} 
\\
&\implies f_i \neq f^{\prime}_j
\end{aligned}$
}
\end{equation}

To separate the rPPG PSDs originating from two different videos, we utilize the negative mean squared error as the loss function for rPPG-rPPG negative pairs, represented as $(f_i,f^{\prime}_j)$. The rPPG-rPPG negative loss term, denoted as $L_n^{RR}$, is presented below, and it is normalized with respect to the total number of rPPG-rPPG negative pairs.

\begin{equation}
L_n^{RR} = - \sum_{i=1}^{N} \sum_{j=1}^{N} \parallel f_i - f^{\prime}_j \parallel^2  / N^2
\label{eq:negative_term}
\end{equation}

\textbf{GT-rPPG Positive Loss.} Inspired by the rPPG temporal similarity, it is expected that the rPPG PSDs from temporal sampling of the ST-rPPG block and the GT PSDs from temporal sampling of the corresponding GT signal should be similar since GT signals are the reference of rPPG signals. The following \textcolor{black}{relations} outline this property. 

For one input video and the corresponding GT signal:
\begin{equation}
\label{eq:tmp_eq}
\resizebox{0.5\textwidth}{!}{
$\begin{aligned}
&\text{PSD}\big\{P(t_1 \to t_1+\Delta t, h_1, w_1)\big\} \approx \text{PSD}\big\{y(t_2 \to t_2+\Delta t)\big\} 
\\
&\implies f_i \approx g_j
\end{aligned}$
}
\end{equation}

For the other video and the corresponding GT signal:
\begin{equation}
\resizebox{0.5\textwidth}{!}{
$\begin{aligned}
&\text{PSD}\big\{P'(t_1 \to t_1+\Delta t, h_1, w_1)\big\} \approx \text{PSD}\big\{y'(t_2 \to t_2+\Delta t)\big\} 
\\
&\implies f'_i \approx g'_j
\end{aligned}$
}
\end{equation}

The GT-rPPG positive loss $L_p^{GR}$ is to pull together rPPG PSDs from one ST-rPPG block and GT PSDs from the corresponding GT signal (GT-rPPG positive pairs, e.g., $(f_i,g_j)$ where $f_i$ is from ST-rPPG block $P$ and $g_j$ is from the corresponding GT signal $y$) so that the model is encouraged to output rPPG signals similar to the corresponding GT signals. Note that this GT-rPPG positive loss does not require exactly synchronized GT signals since rPPG PSDs and GT PSDs are from rPPG samples and GT samples which are randomly temporally sampled from the ST-rPPG block and the GT signal. This indicates that GT-rPPG positive loss does not need the alignment information between the GT signal and the video.  Since it is assumed that some videos may not have GT signals in weakly-supervised learning, the function $\phi$ is defined below to return whether a video has a GT signal.

\begin{equation}
\phi(v) = 
\begin{cases}
      1, & \text{video $v$ has a GT signal}\\
      0, & \text{otherwise}
\end{cases}  
\end{equation}
The GT-rPPG positive loss term $L_p^{GR}$ is defined below, which is normalized by the number of GT-rPPG positive pairs.
\begin{small}
\begin{equation}
L_p^{GR} =  \sum_{i=1}^{N} \sum_{\substack{j=1}}^{N} 
\frac{\phi(v) \parallel f_i - g_j \parallel^2 + \phi(v') \parallel f^{\prime}_i - g^{\prime}_j \parallel^2}{\big( \phi(v)+\phi(v') \big) N^2}
\end{equation}
\end{small}

\textbf{GT-rPPG Negative Loss.} Like the cross-video rPPG dissimilarity, it is expected that the rPPG PSDs sampled from the ST-rPPG block and the GT PSDs from temporal sampling of the non-corresponding GT signal should be different. The following \textcolor{black}{relations} illustrate this property.

For one input video and the non-corresponding GT signal:
\begin{equation}
\resizebox{0.5\textwidth}{!}{
$\begin{aligned}
&\text{PSD}\big\{P(t_1 \to t_1+\Delta t, h_1, w_1)\big\} \neq \text{PSD}\big\{y^{\prime}(t_2 \to t_2+\Delta t)\big\} 
\\
&\implies f_i \neq g^{\prime}_j
\end{aligned}$
}
\end{equation}

For the other input video and the non-corresponding GT signal:
\begin{equation}
\resizebox{0.5\textwidth}{!}{
$\begin{aligned}
&\text{PSD}\big\{P'(t_1 \to t_1+\Delta t, h_1, w_1)\big\} \neq \text{PSD}\big\{y(t_2 \to t_2+\Delta t)\big\} 
\\
&\implies f'_i \neq g_j
\end{aligned}$
}
\end{equation}

The GT-rPPG negative loss term $L_n^{GR}$ pushes away PSDs from one ST-rPPG block and a non-corresponding GT signal (GT-rPPG negative pairs, e.g., $(f_i,g'_j)$ where $f_i$ is from ST-rPPG block $P$ and $g'_j$ is from the non-corresponding GT signal $y'$) so that more negative pairs can be involved during the contrastive learning. \cite{chen2020simple} has demonstrated that more negative samples in contrastive learning can improve performance and facilitate convergence. The GT-rPPG negative loss $L_n^{GR}$ is defined below, which is normalized by the number of GT-rPPG negative pairs.

\begin{small}
\begin{equation}
L_n^{GR} =  -\sum_{i=1}^{N} \sum_{\substack{j=1}}^{N} 
\frac{\phi(v) \parallel f'_i - g_j \parallel^2 + \phi(v') \parallel f_i - g^{\prime}_j \parallel^2}{\big( \phi(v)+\phi(v') \big) N^2}
\end{equation}
\end{small}

\textbf{Overalle Loss.} The overall loss function for Contrast-Phys+ is the combination of the four losses, which can adapt to both unsupervised and weakly-supervised settings.
\begin{equation}
L_{CP+} = L_p^{RR}+L_n^{RR}+L_p^{GR}+L_n^{GR}  
\end{equation}

\subsection{Why Contrast-Phys+ Works with Missing or Unsynchronized Labels} \label{sec: why_cp_plus}

The four rPPG observations are used as constraints to make the model learn the target rPPG signal and exclude noises since noises do not satisfy all observations. Noises that appear in a small local region, such as periodical eye blinking, are excluded since the noises violate rPPG spatial similarity. Noises such as head motions/facial expressions that do not have a temporal constant frequency are excluded since they violate rPPG temporal similarity. The rPPG spatiotemporal similarity is satisfied by minimizing rPPG-rPPG positive loss $L_p^{RR}$. Cross-video rPPG dissimilarity can make two videos' PSDs discriminative and show distinguishable heart rate peaks between two videos' PSDs since heart rate peaks are the main features to distinguish two videos' PSDs as shown in Fig. \ref{fig:cross-video}. Cross-video rPPG dissimilarity is fulfilled by minimizing rPPG-rPPG negative loss $L_n^{RR}$. In addition, PSD values during the heart rate range are used so that noises such as light flickering exceeding the heart rate range are excluded due to the heart rate range constraint.

The loss function $L_{CP+}$ can always be used even though some GT signals are missing. rPPG-rPPG positive loss $L_p^{RR}$ and rPPG-rPPG negative loss $L_n^{RR}$ using rPPG observations do not require GT signals. GT-rPPG positive loss $L_p^{GR}$ and GT-rPPG negative loss $L_n^{GR}$ using GT signals can be adapted to different situations (e.g., Both videos have GT signals, only one video has a GT signal, or neither video has a GT signal).

\begin{figure}[hbt!]
\centering
\begin{minipage}[b]{0.9\linewidth}
  \centering
  \centerline{\includegraphics[width=\linewidth]{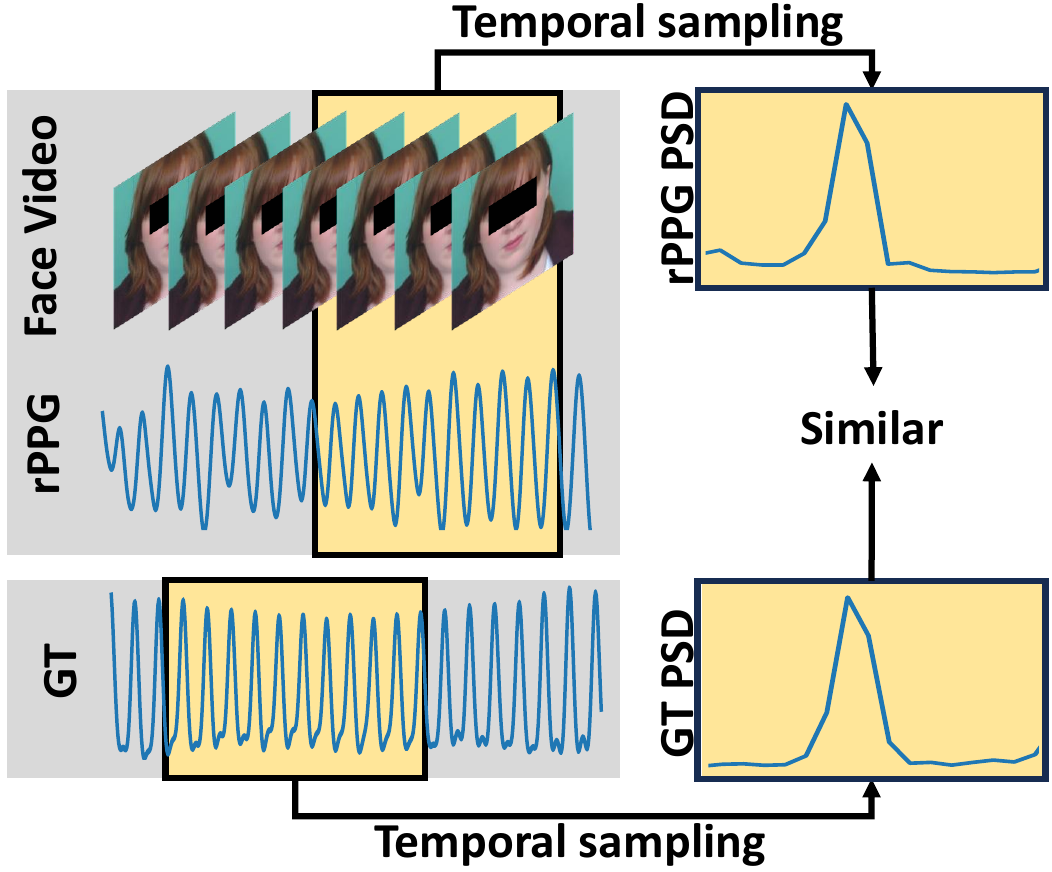}}
\end{minipage}
\caption{Illustration showing that temporally sampled GT/rPPG are similar and independent of the exact synchronization.}
\label{fig:why_unsync}
\end{figure}

Contrast-Phys+ is also robust to unsynchronized GT signals. GT-rPPG negative loss $L_n^{GR}$ is only intended to increase negative pairs using GT samples and rPPG samples for improved contrastive learning \cite{chen2020simple}, so the loss does not require synchronization between facial videos and GT signals. GT-rPPG positive loss $L_p^{GR}$ encourages the rPPG PSD to be similar to the GT PSD. When the GT signal is not precisely synchronized with the facial video, temporally sampled GT/rPPG for the same video can still share similar PSDs since PSDs do not change rapidly in a short time interval as shown in Fig. \ref{fig:why_unsync}. The temporal sampling of GT/rPPG also removes alignment between the GT signal and the video to some extent, making GT-rPPG positive loss $L_p^{GR}$ independent of the exact synchronization. Therefore, temporally sampled GT/rPPG for the same video can be pulled together in the unsynchronized case. We can also use the following \textcolor{black}{relations} to demonstrate that GT-rPPG positive loss $L_p^{GR}$ is robust to GT signal misalignment. Suppose that the GT signal $y(t)$ has a small misalignment $u$, resulting $y(t+u)$. The PSDs of temporal samples of $y(t)$ and $y(t+u)$ are $\text{PSD}\{y(t_2 \to t_2+\Delta t)\}$ and $\text{PSD}\{y(t_2+u \to t_2+u+\Delta t)\}$, respectively. According to the temporal similarity in Sec. \ref{sec:temp_similarity},

\begin{equation}
\text{PSD}\{y(t_2 \to t_2+\Delta t)\} \approx \text{PSD}\{y(t_2+u \to t_2+u+\Delta t)\}
\end{equation}
holds if $|t_2+u-t_2|=|u|$ is small where $u$ is the small misalignment. Combine the above \textcolor{black}{relation} with \textcolor{black}{relation} \ref{eq:tmp_eq}, we get

\begin{equation}
\resizebox{0.5\textwidth}{!}{
$\begin{aligned}
\text{PSD}\big\{P(t_1 \to t_1+\Delta t, h_1, w_1)\big\} &\approx \text{PSD}\big\{y(t_2 \to t_2+\Delta t)\big\} 
\\
&\approx \text{PSD}\{y(t_2+u \to t_2+u+\Delta t)\}
\end{aligned}$
}
\end{equation}
which indicates that rPPG samples from the ST-rPPG block $P$ are similar to the GT samples from the misaligned GT signal $y(t+u)$. Therefore, our method is robust to GT signal misalignment.

\section{Experiments}
% This section will show our experimental results, including intra-dataset testing, cross-dataset testing, running speed test, saliency map visualization, and sensitivity analysis.

\subsection{Experimental Setup and Metrics}

\subsubsection{Datasets}

We conducted experiments using five common rPPG datasets, encompassing RGB and NIR videos recorded under diverse scenarios. Specifically, we employed the PURE dataset \cite{stricker2014non}, UBFC-rPPG dataset \cite{bobbia2019unsupervised}, OBF dataset \cite{li2018obf}, and MR-NIRP dataset \cite{nowara2018sparseppg,nowara2020near} for intra-dataset evaluations. Additionally, we employed the MMSE-HR dataset \cite{zhang2016multimodal} for both intra-dataset and cross-dataset evaluations. We also use the EquiPleth dataset \cite{vilesov2022blending} to evaluate the fairness performance on different skin tones. In addition, we also compare our method with other baselines when training on synthetic rPPG datasets including UCLA-Synth \cite{wang2022synthetic} and SCAMPS \cite{mcduff2022scamps}.

\textbf{PURE} \cite{stricker2014non} comprises facial videos from ten subjects recorded in six distinct setups, encompassing both static and dynamic tasks. To ensure consistency, we followed the same experimental protocol used in previous studies \cite{vspetlik2018visual,lu2021dual} for partitioning the training and test sets. 

\textbf{UBFC-rPPG} \cite{bobbia2019unsupervised} comprises facial videos from 42 subjects who participated in a mathematical game designed to elevate their heart rates. For evaluation, we adhered to the protocol outlined in \cite{lu2021dual} for train-test split. 

\textbf{OBF} \cite{li2018obf} encompasses 200 videos from 100 healthy subjects recorded both before and after exercise sessions. To facilitate a fair comparison with prior work, we conducted subject-independent ten-fold cross-validation, as previously described in \cite{yu2019remoteBMVC,yu2019remote,niu2020video}. 

\textbf{MR-NIRP} \cite{nowara2018sparseppg,nowara2020near} contains NIR videos of eight subjects, capturing instances of subjects remaining stationary as well as engaging in motion tasks. Due to its limited scale and the inherent challenge of weak rPPG signals in NIR \cite{martinez2011optimal,vizbara2013comparison}, we employed a leave-one-subject-out cross-validation protocol for our experiments. Both stationary and motion videos are used.

\textbf{MMSE-HR} \cite{zhang2016multimodal} includes 102 videos from 40 subjects recorded during emotion elicitation experiments. Given the presence of spontaneous facial expressions and head movements, we conducted subject-independent 5-fold cross-validation for intra-dataset testing on the MMSE-HR dataset. Further details regarding these datasets are available in the supplementary material.

\textbf{EquiPleth} \cite{vilesov2022blending} includes 91 participants (28 light. 49 medium, and 14 dark skin tone participants). Each subject has 6 recordings, and each recording has a 30s RGB video and radar data. The GT PPG signals are recorded and synchronized with the RGB videos and radar data. We use the predefined train, validation, and test splits and report the results on the test set. We use the RGB data and GT PPG signals to train and test our method on different skin tones.

There are two synthetic rPPG datasets including \textbf{UCLA-Synth} \cite{wang2022synthetic} and \textbf{SCAMPS} \cite{mcduff2022scamps}. The avatar facial videos are rendered by graphic pipelines. The GT PPG signals are temporally modulated with skin albedo maps to produce the skin color variations induced by rPPG. In addition, head motions and facial expressions are added during rendering. UCLA-Synth has 476 avatar subjects including 120 African, 118 Asian, 120 Caucasian, and 118 Indian avatar subjects. Each avatar subject has a 70-second facial video and GT signal. SCAMPS has 2800 avatar subjects with diverse skin tones and facial appearances. Each avatar subject has a 20-second facial video and GT signal.

\subsubsection{Experimental Setup} 
During each training iteration, the model receives two 10-second clips from two different videos as inputs. If available, ground truth (GT) signals are incorporated; for instance, 20\% of the videos contain GT signals, or in some cases, all videos possess unsynchronized GT signals. To train Contrast-Phys+ effectively, we employ the AdamW optimizer \cite{loshchilov2018decoupled} with a learning rate of $10^{-5}$, training the model for 30 epochs on a single NVIDIA Tesla V100 GPU. Following the approach in \cite{gideon2021way}, we select the model with the lowest irrelevant power ratio (IPR) on the training set to achieve model selection (for further insights into IPR, refer to the supplementary materials).

During the testing phase, we segment each test video into non-overlapping 30-second clips and extract the rPPG signal from each clip. To compute the heart rate (HR), we identify the HR peak in the PSD of the rPPG signal. Additionally, we employ Neurokit2 \cite{makowski2021neurokit2} to locate systolic peaks within the rPPG signals, allowing us to derive heart rate variability (HRV) metrics.

According to our ablation study for the ST-rPPG blocks in Sec. \ref{sec:ablation_st_rppg_block}, we set the spatial resolution of the ST-rPPG block to be $2 \times 2$, with the time duration of 10 seconds. For the rPPG spatiotemporal sampling process, we use $K=4$, indicating that, for each spatial position within the ST-rPPG block, four rPPG samples are randomly selected. The time interval $\Delta t$ between each rPPG sample is set to 5 seconds, which is half of the time duration of the ST-rPPG block. Consequently, we obtain 16 rPPG samples ($N=16$) from each ST-rPPG block. Regarding the temporal sampling of GT signals, we maintain the same $\Delta t$ of 5 seconds, resulting in the selection of 16 GT samples ($N=16$) from a GT signal.

\subsubsection{Evaluation Metrics}

In line with prior research \cite{li2014remote,niu2020video,yu2019remote}, we use three metrics to assess the accuracy of heart rate (HR) measurement: the mean absolute error (MAE), root mean squared error (RMSE), and Pearson correlation coefficient (R). Additionally, we utilize the signal-to-noise ratio (SNR) \cite{de2013robust} to evaluate the quality of the rPPG signal. For the evaluation of HRV features, which encompass respiration frequency (RF), low-frequency power (LF) in normalized units (n.u.), high-frequency power (HF) in normalized units (n.u.), and the LF/HF power ratio, we follow the approach outlined in \cite{lu2021dual} and employ the standard deviation (STD), RMSE, and R as evaluation metrics. In the context of MAE, RMSE, and STD, smaller values indicate lower errors, whereas for R, higher values approaching one denote reduced errors. For SNR, larger values indicate higher-quality rPPG signals. For a more comprehensive understanding of these evaluation metrics, please refer to the supplementary material.

\subsection{Intra-dataset Testing}

\subsubsection{HR Estimation}

We conducted intra-dataset testing for HR estimation on four datasets: PURE, UBFC-rPPG, OBF, and MR-NIRP. Contrast-Phys+ was trained under various conditions, including scenarios where 0\%, 20\%, or 60\% of the videos contain GT signals. These settings represent the unsupervised and semi-supervised paradigms, with the semi-supervised setup encompassing partially available labels. Additionally, Contrast-Phys+ was trained with 100\% of the labels, representing the supervised setting.

The results of HR estimation for Contrast-Phys+ are presented in Table \ref{tb:intra} and compared against multiple baseline methods. These baselines include traditional methods, supervised methods, semi-supervised methods, and recent unsupervised methods. Notably, Contrast-Phys+ (0\%) outperforms several unsupervised baselines \cite{gideon2021way,speth2023non,yue2023facial} and comes remarkably close to the performance of supervised methods \cite{lu2021dual,niu2020video,nowara2021benefit}. In the semi-supervised setting, when partial GT signals are available (Contrast-Phys+ 20\% and 60\%), the performance improves further, often surpassing recent supervised methods \cite{lu2021dual,niu2020video,nowara2021benefit}. In the supervised setting (Contrast-Phys+ (100\%)), Contrast-Phys+ achieves the best performance among supervised methods across most evaluation metrics. This underscores the advantage of Contrast+Phys+ as it learns from both labels and rPPG observations, whereas previous supervised methods only rely on labels. The consistently superior performance of Contrast-Phys+ holds across all four datasets, including the MR-NIRP dataset containing NIR videos.

\begin{table*}[htb!]
\caption{Intra-dataset HR results. The best results are in bold, and the second-best results are underlined.}
% \fontsize{6.5}{8}\selectfont
% \setlength{\tabcolsep}{1.5pt}
\resizebox{\linewidth}{!}
{
\begin{tabular}{llcccccccccccc} 
\toprule
\multirow{2}{*}{\begin{tabular}[c]{@{}l@{}}\\Method\\ Types\end{tabular}}                                                                                                                   & \multirow{2}{*}{\begin{tabular}[c]{@{}l@{}}\\Methods\end{tabular}}  & \multicolumn{3}{c}{UBFC-rPPG}                                                                                                     & \multicolumn{3}{c}{PURE}                                                                                                          & \multicolumn{3}{c}{OBF}                                                                                                           & \multicolumn{3}{c}{MR-NIRP (NIR)}                                                                                                  \\ 
\cmidrule(lr){3-5}\cmidrule(lr){6-8}\cmidrule(lr){9-11}\cmidrule(lr){12-14}
& & \begin{tabular}[c]{@{}c@{}}MAE\\ (bpm)\end{tabular} & \begin{tabular}[c]{@{}c@{}}RMSE\\ (bpm)\end{tabular} & R                    & \begin{tabular}[c]{@{}c@{}}MAE\\ (bpm)\end{tabular} & \begin{tabular}[c]{@{}c@{}}RMSE\\ (bpm)\end{tabular} & R                    & \begin{tabular}[c]{@{}c@{}}MAE\\ (bpm)\end{tabular} & \begin{tabular}[c]{@{}c@{}}RMSE\\ (bpm)\end{tabular} & R                    & \begin{tabular}[c]{@{}c@{}}MAE\\ (bpm)\end{tabular} & \begin{tabular}[c]{@{}c@{}}RMSE\\ (bpm)\end{tabular} & R                     \\ 
\midrule
\multirow{5}{*}{\begin{tabular}[c]{@{}l@{}}Tradi-\\tional\end{tabular}}                                          & GREEN \cite{verkruysse2008remote} & 7.50 & 14.41 & 0.62 & - & - & - & - & 2.162 & 0.99 & - & - & - \\
& ICA \cite{poh2010advancements} & 5.17 & 11.76 & 0.65 & - & - & - & - & - & - & - & - & - \\
& CHROM \cite{de2013robust} & 2.37 & 4.91 & 0.89 & 2.07 & 9.92 & \underline{0.99} & - & 2.733 & 0.98 & - & - & - \\
& 2SR \cite{wang2015novel} & - & - & - & 2.44 & 3.06 & 0.98 & - & - & - & - & - & -\\
& POS \cite{wang2016algorithmic} & 4.05 & 8.75 & 0.78 & - & - & - & - & 1.906 & 0.991 & - & - & - \\
 
\midrule
\multirow{10}{*}{\begin{tabular}[c]{@{}l@{}}Super-\\vised\end{tabular}} 

& CAN \cite{chen2018deepphys} & - & - & - & - & - & - & - & - & - & 7.78 & 16.8 & -0.03\\
& HR-CNN \cite{vspetlik2018visual} & - & - & - & 1.84 & 2.37 & 0.98 & - & - & - & - & - & -\\
& SynRhythm \cite{niu2018synrhythm} & 5.59 & 6.82 & 0.72 & - & - & - & - & - & - & - & - & - \\
& PhysNet \cite{yu2019remoteBMVC} & - & - & - & 2.1 & 2.6 & \underline{0.99} & - & 1.812 & 0.992 & 3.07 & 7.55 & 0.655\\
& rPPGNet \cite{yu2019remote} & - & - & - & - & - & - & - & 1.8 & 0.992 & - & - & -\\
& CVD \cite{niu2020video} & - & - & - & - & - & - & - & 1.26 & 0.996 & - & - & -\\
& PulseGAN \cite{song2021pulsegan} & 1.19 & 2.10 & \underline{0.98} & - & - & - & - & - & - & - & - & - \\
& Dual-GAN \cite{lu2021dual} & 0.44 & \bf{0.67} & \bf{0.99} & 0.82 & 1.31 & \underline{0.99} & - & - & - & - & - & - \\
& Nowara2021 \cite{nowara2021benefit} & - & - & - & - & - & - & - & - & - & \underline{2.34} & 4.46 & 0.85\\
& \bf Contrast-Phys+ (100\%) & \bf 0.21 & \underline{0.80} & \bf 0.99 & \bf 0.48 & \bf 0.98 & \underline{0.99} & \bf 0.34 & \bf 0.75 & \bf 0.998 & \bf 1.96 & \bf 3.02 & \bf 0.93 \\ 
\midrule
\multirow{2}{*}{\begin{tabular}[c]{@{}l@{}}Semi-su-\\pervised\end{tabular}}
& \bf Contrast-Phys+ (60\%) & \underline{0.22} & 0.81 & \bf 0.99 & \underline{0.52} & \underline{1.02} & \underline{0.99} & \underline{0.35} & \underline{0.79} & \underline{0.997} & 2.58 & \underline{3.65} & \underline{0.89} \\ 
& \bf Contrast-Phys+ (20\%) & 0.24 & 0.87 & \bf 0.99 & 0.61 & 1.18 & \underline{0.99} & 0.37 & 0.84 & \underline{0.997} & 2.57 & 4.02 & 0.88\\ 
\midrule
\multirow{4}{*}{\begin{tabular}[c]{@{}l@{}}Unsuper-\\vised\end{tabular}}
& \bf Contrast-Phys+ (0\%) & \begin{tabular}[c]{@{}l@{}} 0.64 \end{tabular} & \begin{tabular}[c]{@{}l@{}} 1.00 \end{tabular} & \begin{tabular}[c]{@{}l@{}} \bf{0.99} \end{tabular} & 1.00 & 1.40 & \underline{0.99} & 0.51 & 1.39 & 0.994 & 2.68 & 4.77 & 0.85 \\
& Gideon2021 \cite{gideon2021way} & \begin{tabular}[c]{@{}l@{}}1.85 \end{tabular} & \begin{tabular}[c]{@{}l@{}}4.28 \end{tabular} & \begin{tabular}[c]{@{}l@{}}0.93 \end{tabular}   & 2.3 & 2.9 & \underline{0.99} & 2.83 & 7.88 & 0.825 & 4.75 & 9.14 & 0.61\\
& SiNC \cite{speth2023non} & 0.59 & 1.83 & \bf 0.99 & 0.61 & 1.84 & \bf 1.00 & - & - & - & - & - & - \\
& Yue \textit{et al.}\cite{yue2023facial} & 0.58 & 0.94 & \bf 0.99 & 1.23 & 2.01 & \underline{0.99} & - & - & - & - & - & - \\
\bottomrule
\end{tabular}}
\label{tb:intra}
\end{table*}

\subsubsection{HRV Estimation}

\begin{table*}[htb!]
\caption{HRV results on UBFC-rPPG. The best results are in bold, and the second-best results are underlined.}
\centering
% \fontsize{6.5}{8}\selectfont
% \setlength{\tabcolsep}{2pt}
\resizebox{\linewidth}{!}{
\begin{tabular}{llcccccccccccc} 
\toprule
\multirow{2}{*}{\begin{tabular}[c]{@{}l@{}} Method\\ Types\end{tabular}} & \multirow{2}{*}{\begin{tabular}[c]{@{}l@{}} Methods\end{tabular}}  & \multicolumn{3}{c}{LF (n.u.)}               & \multicolumn{3}{c}{HF (n.u.)} & \multicolumn{3}{c}{LF/HF} & \multicolumn{3}{c}{RF(Hz)} \\ 
\cmidrule(lr){3-5}\cmidrule(lr){6-8}\cmidrule(lr){9-11}\cmidrule(lr){12-14}&                          
& STD & RMSE & R & STD & RMSE & R & STD & RMSE & R & STD & RMSE & R
\\ 
\midrule
\multirow{3}{*}{\begin{tabular}[c]{@{}l@{}}Tradi-\\tional\end{tabular}}
& GREEN \cite{verkruysse2008remote} & 0.186 & 0.186 & 0.280 & 0.186 & 0.186 & 0.280 & 0.361 & 0.365 & 0.492 & 0.087 & 0.086 & 0.111 \\
& ICA \cite{poh2010advancements} & 0.243 & 0.240 & 0.159 & 0.243 & 0.240 & 0.159 & 0.655 & 0.645 & 0.226 & 0.086 & 0.089 & 0.102 \\
& POS \cite{wang2016algorithmic} & 0.171 & 0.169 & 0.479 & 0.171 & 0.169 & 0.479 & 0.405 & 0.399 & 0.518 & 0.109 & 0.107 & 0.087 \\
\midrule
\multirow{3}{*}{\begin{tabular}[c]{@{}l@{}}Super-\\vised\end{tabular}} 
& CVD \cite{niu2020video} & 0.053 & 0.065 & 0.740 & 0.053 & 0.065 & 0.740 & 0.169 & 0.168 & 0.812 & \underline{0.017} & \underline{0.018} & 0.252\\
& Dual-GAN \cite{lu2021dual} & 0.034 & 0.035 & 0.891 & 0.034 & 0.035 & 0.891 & 0.131 & 0.136 & 0.881 & \bf{0.010} & \bf{0.010} & 0.395 \\
& \bf Contrast-Phys+ (100\%) & \bf 0.025 & \bf 0.025 & \bf 0.947 & \bf 0.025 & \bf 0.025 & \bf 0.947 & \bf 0.064 & \bf 0.066 & \bf 0.963 & 0.029 & 0.029 & \bf 0.803  \\
\midrule
\multirow{2}{*}{\begin{tabular}[c]{@{}l@{}}Semi-su-\\pervised\end{tabular}} 
& \bf Contrast-Phys+ (60\%) & \underline{0.035} & \underline{0.035} & \underline{0.908} & \underline{0.035} & \underline{0.035} & \underline{0.908} & \underline{0.100} & \underline{0.105} & \underline{0.906} & 0.036 & 0.043 & \underline{0.746}\\
& \bf Contrast-Phys+ (20\%) & 0.037 & 0.037 & 0.882 & 0.037 & 0.037 & 0.882 & 0.119 & 0.120 & 0.866 & 0.037 & 0.040 & 0.662 \\
\midrule
\multirow{2}{*}{\begin{tabular}[c]{@{}l@{}}Unsup-\\ervised\end{tabular}}
& \bf Contrast-Phys+ (0\%) & 0.096 & 0.098 & 0.798 & 0.096 & 0.098 & 0.798 & 0.391 & 0.395 & 0.782 & 0.085 & 0.083 &  0.347 \\
& Gideon2021 \cite{gideon2021way} & 0.142 & 0.139 & 0.694 & 0.142 & 0.139 & 0.694 & 0.687 & 0.691 & 0.684 & 0.098 & 0.098 & 0.103 \\

\bottomrule
\end{tabular}}
\label{tb:HRV}
\end{table*}

Intra-dataset testing for heart rate variability (HRV) evaluation was conducted on the UBFC-rPPG dataset, and the results are presented in Table \ref{tb:HRV}. HRV analysis demands precisely measured, high-quality rPPG signals for accurate systolic peak detection. Notably, Contrast-Phys+ significantly outperforms traditional methods and the previous unsupervised baseline \cite{gideon2021way} in terms of HRV results. When partial GT signals are incorporated, the performance of Contrast-Phys+ closely approaches that of supervised methods. In the case of Contrast-Phys+ utilizing all labels (100\%), it achieves the best results across most HRV metrics. These findings underscore the capability of Contrast-Phys+ to yield high-quality rPPG signals with accurate systolic peaks, enabling the derivation of HRV features. This feature makes it a promising candidate for applications in emotion understanding \cite{yu2021facial,mcduff2014remote,sabour2021ubfc} and healthcare \cite{shi2019atrial,yan2018contact}. Additionally, Contrast-Phys+ has the potential to further refine its understanding of rPPG signals by leveraging GT information, as illustrated in Section \ref{sec:rppg_waveform}.

\subsection{Cross-dataset Testing}

\begin{table*}[htb!]
\caption{Cross-dataset and intra-dataset HR results for MMSE-HR. The best results are in bold, and the second-best results are underlined.}
% \fontsize{6.5}{8}\selectfont
% \setlength{\tabcolsep}{1.5pt}
\resizebox{\linewidth}{!}{
\begin{tabular}{llcccccccc} 
\toprule
\multirow{2}{*}{\begin{tabular}[c]{@{}l@{}}\\Method Types\end{tabular}}                                                                                        & \multirow{2}{*}{\begin{tabular}[c]{@{}l@{}}\\Methods\end{tabular}}  & \multicolumn{4}{c}{Cross-dataset (UBFC $\rightarrow$ MMSE-HR)}& \multicolumn{4}{c}{Intra-dataset (MMSE-HR $\rightarrow$ MMSE-HR)} \\ 
\cmidrule(lr){3-6}\cmidrule(lr){7-10}
& & \begin{tabular}[c]{@{}c@{}}MAE (bpm)\end{tabular} & \begin{tabular}[c]{@{}c@{}}RMSE (bpm)\end{tabular} & R & SNR (dB) & \begin{tabular}[c]{@{}c@{}}MAE (bpm)\end{tabular} & \begin{tabular}[c]{@{}c@{}}RMSE (bpm)\end{tabular} & R & SNR (dB) \\ 
\midrule
\multirow{3}{*}{\begin{tabular}[c]{@{}l@{}}Traditional\end{tabular}} 
& Li2014 \cite{poh2010advancements} & - & 19.95 & 0.38 & - & - & 19.95 & 0.38 & - \\
& CHROM \cite{de2013robust} & - & 13.97 & 0.55 & - & - & 13.97 & 0.55 & -  \\
& SAMC \cite{tulyakov2016self} & - & 11.37 & 0.71 & - & - & 11.37 & 0.71 & - \\
\midrule
\multirow{4}{*}{\begin{tabular}[c]{@{}l@{}}Supervised\end{tabular}} 

& PhysNet \cite{yu2019remoteBMVC} & \underline{2.04} & 6.85 & 0.86 & 1.17 & 1.22 & 4.49 & 0.94 & 2.8  \\
& TS-CAN \cite{NEURIPS2020_e1228be4} & 3.41 & 9.29 & 0.76 & -1.18 & 2.89 & 7.18 & 0.86 & -2.01 \\
& PhysFormer \cite{yu2022physformer} & 2.68 & 7.01 & 0.86 & 1.2 & 1.48 & 4.22 & \underline{0.95} & 2.55 \\
& \bf Contrast-Phys+ (100\%) & \bf 1.76 & \bf 5.34 & \bf 0.92 & \bf 1.37 & \bf 1.11 & \bf 3.83 & \bf 0.96 & \bf 3.72 \\ 
\midrule
\multirow{2}{*}{\begin{tabular}[c]{@{}l@{}}Semi-supervised\end{tabular}}
& \bf Contrast-Phys+ (60\%) & 2.30 & \underline{6.32} & \underline{0.89} & \underline{1.25} & \underline{1.20} & \underline{3.89} & \bf 0.96 & \underline{3.51} \\ 
& \bf Contrast-Phys+ (20\%) & 2.28 & 6.51 & 0.88 & 1.15 & 1.51 & 4.15 & \underline{0.95} & 2.93 \\ 
\midrule
\multirow{2}{*}{\begin{tabular}[c]{@{}l@{}}Unsupervised\end{tabular}}
& \bf Contrast-Phys+ (0\%) & 2.43 & 7.34 & 0.86 & 1.09 & 1.82 & 6.69 & 0.87 & 2.64 \\
& Gideon2021 \cite{gideon2021way} & 4.10 & 11.55 & 0.70 & 0.26 & 3.98 & 9.65 & 0.85 & 0.67 \\
\bottomrule
\end{tabular}}
\label{tb:cross}
\end{table*}

\begin{table}[]
\caption{Cross-dataset results of Contrast-Phys+ when additional unlabeled videos (PURE and OBF) are used for training. The best results are in bold.}
% \fontsize{7.8}{10}\selectfont
% \setlength{\tabcolsep}{2.3pt}
\resizebox{\linewidth}{!}{
\begin{tabular}{@{}c|cc|cccc@{}}
\hline
\multicolumn{3}{c}{Training Sets} & \multicolumn{4}{|c}{\begin{tabular}[c]{@{}c@{}}Cross-dataset Results\\ (test on MMSE-HR)\end{tabular}}\\ 
% \cmidrule(lr){1-3}\cmidrule(lr){4-7}
\hline
\begin{tabular}[c]{@{}c@{}} Labeled\\ UBFC\end{tabular} & \begin{tabular}[c]{@{}c@{}}Unlabeled\\ PURE\end{tabular} & \begin{tabular}[c]{@{}c@{}}Unlabeled\\ OBF\end{tabular} & \begin{tabular}[c]{@{}c@{}}MAE\\ (BPM)\end{tabular} & \begin{tabular}[c]{@{}c@{}}RMSE\\ (bpm)\end{tabular} & R & \begin{tabular}[c]{@{}c@{}}SNR\\ (dB)\end{tabular} \\ 
\hline
\checkmark &  & & 1.76 & 5.34 & 0.92 & 1.37 \\
\hline
\checkmark & \checkmark & & 1.13 & \bf 3.71 & \bf 0.96 & 2.37  \\
\checkmark &   & \checkmark & 1.47 & 4.55 & 0.94 & \bf 3.17 \\
\hline
\end{tabular}
}
\label{tb:cross_with_other}
\end{table}

We perform cross-dataset testing on MMSE-HR to test the generalization of the proposed methods. We train recent supervised methods \cite{yu2019remoteBMVC,NEURIPS2020_e1228be4,yu2022physformer,yu2023physformer++}, the unsupervised baseline \cite{gideon2021way}, and Contrast-Phys+ on UBFC and test the models on MMSE-HR. In addition, we also provide intra-dataset results by training and testing the models on MMSE-HR as a reference to be compared with the cross-dataset results. Tab. \ref{tb:cross} shows the cross-dataset and intra-dataset results on MMSE-HR, which can be summarized in four aspects as below. \textbf{1)} First, Contrast-Phys+ achieves good cross-dataset results compared with other supervised and unsupervised baselines, which means the proposed method can generalize well to a new dataset. The results are very promising, as in practical applications, we might potentially use enormous facial videos from different sources with no/partial GT signals to train Contrast-Phys/Contrast-Phys+ and then apply them to the target data. \textbf{2)} Second, more labels from Contrast-Phys+ (0\%, unsupervised) to Contrast-Phys+ (100\%, fully supervised) can provide better performance for both cross- and intra-dataset results, which means additional GT signals can help fit rPPG signals and improve generalization. \textbf{3)} Third, for both cross- and intra-dataset results, Contrast-Phys+ (100\%) using both label information and rPPG observations achieves better performance than other supervised methods that only utilize label information. Therefore, rPPG observations as the prior knowledge play an important role in improving rPPG measurement performance in the fully supervised setting. \textbf{4)} Last, comparing cross- and intra-dataset results, performance for intra-dataset is generally better than for cross-dataset for each deep learning-based method, so training and testing on the same dataset are preferred to keep good performance. Compared with previous supervised methods, Contrast-Phys+ lowers the requirement of intra-dataset training since it only needs facial videos with no or partial labels.

Contrast-Phys+ exhibits the capability to adapt to both labeled and unlabeled videos during training, allowing for the expansion and diversification of the training dataset by incorporating unlabeled videos from other sources. This augmentation strategy aims to enhance the model's generalization. To this end, we employed all labeled UBFC videos alongside additional unlabeled videos from PURE or OBF to train Contrast-Phys+ and evaluated the model's performance on MMSE-HR.

The results in Table \ref{tb:cross_with_other} demonstrate that the inclusion of additional unlabeled videos for training results in improved performance compared to training solely with labeled UBFC data. When additional unlabeled training data is introduced, the cross-dataset testing performance even approaches the levels achieved by the best intra-dataset testing performance, as demonstrated in Table \ref{tb:cross}. This suggests that Contrast-Phys+ can seamlessly expand its training dataset by incorporating unlabeled videos from different domains, thereby enhancing generalization and achieving performance levels close to intra-dataset results. Such a capability was not feasible with previous supervised methods, highlighting the strengths of Contrast-Phys+.

\subsection{Training with Unsynchronized GT Signals}

\begin{figure}[hbt!]
\centering
\begin{minipage}[b]{0.49\linewidth}
  \centering
  \centerline{\includegraphics[width=\linewidth]{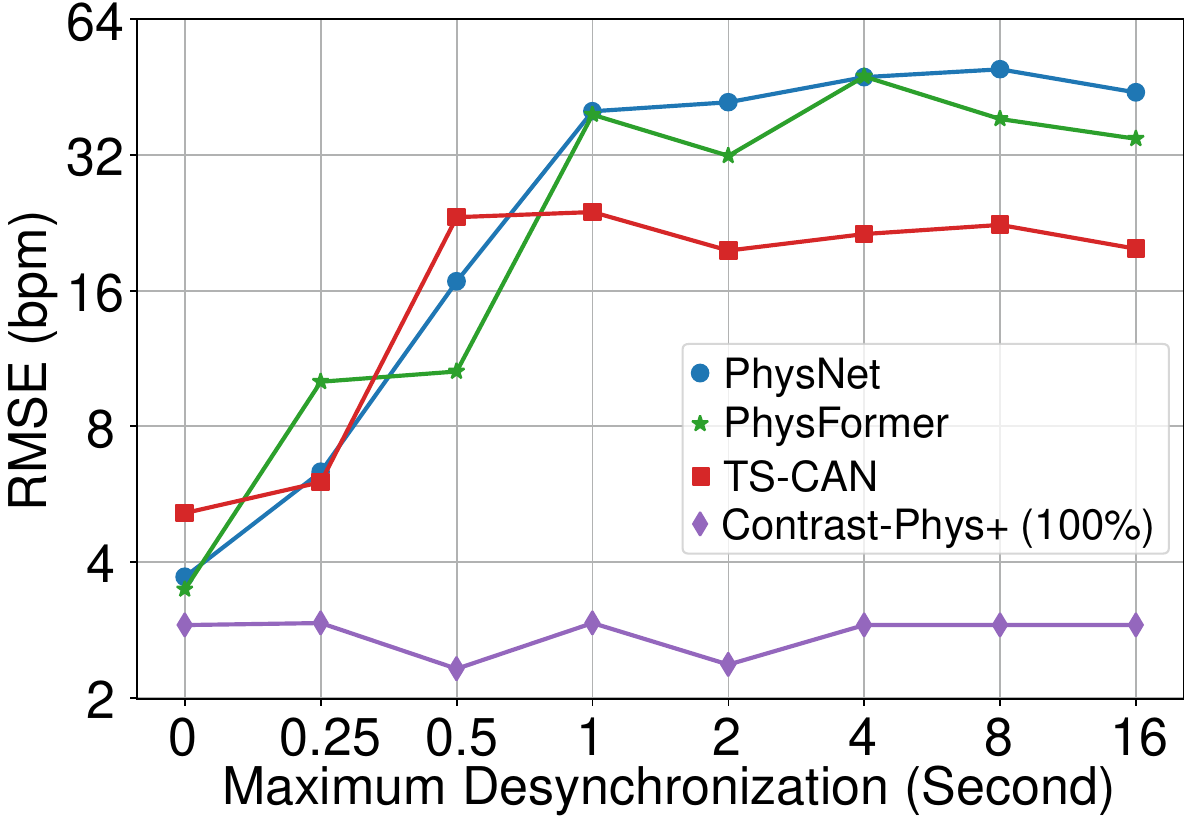}}
  \centerline{\footnotesize (a) RMSE}
\end{minipage}
\begin{minipage}[b]{0.49\linewidth}
  \centering
  \centerline{\includegraphics[width=\linewidth]{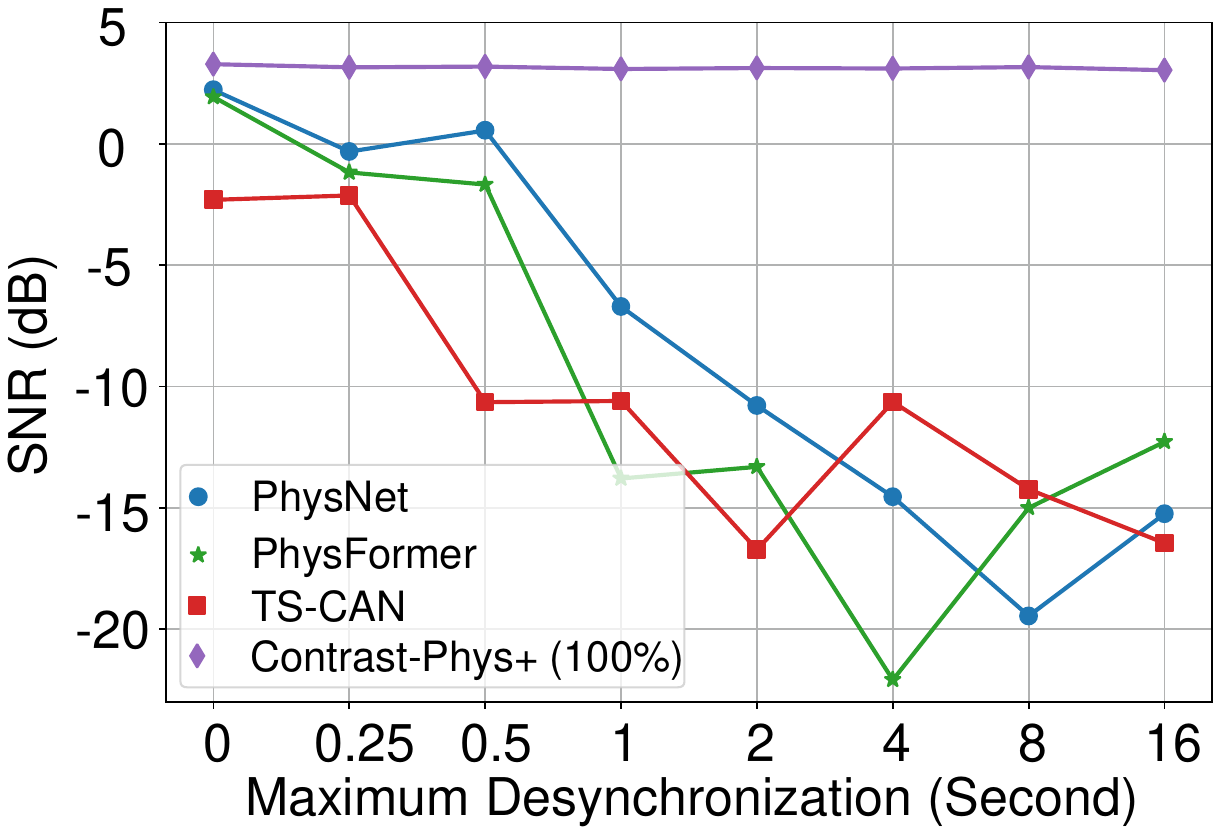}}
  \centerline{\footnotesize (b) SNR}
\end{minipage}
\caption{rPPG measurement performance ((a) RMSE and (b) SNR) with respect to maximum desynchronization of GT signals.}
\label{fig:desync}
\end{figure}

In our experiments, we explored scenarios where ground truth (GT) signals are desynchronized from the facial videos, which is a noisy label case in weakly-supervised learning. We introduced a parameter, the maximum desynchronization $D_{\text{max}}$, and temporally shifted each GT signal by a random offset within the range of $-D_{\text{max}}$ to $D_{\text{max}}$, ensuring that the GT signals were no longer synchronized with the corresponding facial videos. This desynchronization was applied to GT signals in the training set, after which we trained the model and evaluated its performance on the test set. This experiment was tested on a single fold of the MMSE-HR dataset and trained on the other 4 folds.

As depicted in Fig. \ref{fig:desync}, we analyzed the RMSE and SNR across various levels of maximum desynchronization. Notably, as the maximum desynchronization increased, the performance of previous supervised methods exhibited significant deterioration. Even a small maximum desynchronization of 0.25 seconds, which is realistic and likely to occur during data collection, considerably impacted their performance.

In contrast, Contrast-Phys+ (100\%) demonstrated robust and stable performance in terms of both RMSE and SNR across different maximum desynchronization values. These results underscore the robustness of Contrast-Phys+ to GT signal desynchronization, while previous supervised methods proved to be highly susceptible to even minor misalignments. This robustness can be attributed to Contrast-Phys+'s use of PSD instead of pulse curves in the temporal domain, which is comparatively stable over short time intervals. Consequently, learning an rPPG signal with misaligned GT signals in the frequency domain, aided by the rPPG observation constraint, is a viable approach as demonstrated in Sec. \ref{sec: why_cp_plus}. These results indicate that Contrast-Phys+ offers greater tolerance when facial videos and GT signals are not perfectly aligned, streamlining the rPPG data collection process.

\subsection{Evaluation of Skin Tone Fairness}

\begin{table}[htb!]
\centering
\caption{HR performance and fairness on EquiPleth dataset. RMSE measures the HR performance on the entire dataset. Fairness is the RMSE difference between dark and light skin groups (Lower values mean better fairness.). The best results are in bold, and the second-best results are underlined.}
\label{tab:fairness}
\begin{tabular}{@{}llcc@{}}
\toprule
Modalities             & Methods                & \begin{tabular}[c]{@{}c@{}}RMSE $\downarrow$ \\ (bpm)\end{tabular} & \begin{tabular}[c]{@{}c@{}}Fairness$ \downarrow$ \\ (bpm)\end{tabular} \\ \midrule
\multirow{4}{*}{RGB} & PhysNet \cite{yu2019remoteBMVC}                & 5.26       & 4.05           \\
                     & Contrast-Phys+ (0\%)   & 5.16       & 3.39           \\
                     & Contrast-Phys+ (50\%)  & 4.97       & 3.05           \\
                     & Contrast-Phys+ (100\%) & \underline{4.40}       & 1.96           \\ \midrule
Radar                & Vilesov et al. \cite{vilesov2022blending}         & 6.12       & \bf 0.85           \\ \midrule
RGB+Radar            & Vilesov et al. \cite{vilesov2022blending}         & \bf 3.42       & \underline{1.44}           \\ \bottomrule
\end{tabular}
\end{table}

Previous studies \cite{kadambi2021achieving,vilesov2022blending,nowara2020meta} have pointed out that rPPG algorithms based on RGB videos have performance bias in different skin tones. RGB video-based measurement is less accurate on subjects with darker skin, since darker skin absorbs more light resulting in weaker skin color changes. We test our method on the EquiPleth dataset \cite{vilesov2022blending} to study the performance gap between dark and light skin tones. We also add baselines from Vilesov et al. \cite{vilesov2022blending} that utilized radar and RGB+radar to improve skin tone fairness and overall performance. Table \ref{tab:fairness} shows the overall performance and fairness on the EquiPleth dataset. For the overall RMSE performance, RGB+Radar fusion achieves the best performance, and Contrast-phys+ (100\%) achieves the second best. The results indicate that RGB+Radar multimodal fusion can improve the overall performance. For fairness (the RMSE difference between dark and light skin groups), radar and radar+RGB achieve the best and second-best fairness since radar detects chest movement for heart rate measurement and does not depend on skin color. The results agree with the previous finding \cite{vilesov2022blending} that, in general, RGB video-based methods show larger skin tone bias than the radar approach. On the other hand, among RGB methods, Contrast-Phys+ achieves better fairness than PhysNet, which means we can develop new video-based rPPG algorithms to improve skin tone fairness.

\subsection{Training with Synthetic rPPG Data}

To solve the GT lacking issue, one solution is to develop unsupervised or semi-supervised methods like Contrast-Phys+, and another is to train supervised models on synthetic rPPG data \cite{wang2022synthetic,mcduff2022scamps}. We take the real UBFC-rPPG dataset as the test set, and compare the two solutions in three settings: 1) previous supervised methods and supervised Contrast-Phys+ trained on labeled synthetic data, 2) semi-supervised Contrast-Phys+ trained on labeled synthetic data and unlabeled real data (unlabeled MMSE-HR), 3) unsupervised Contrast-Phys+ trained on unlabeled real data (unlabeled MMSE-HR). 

\begin{table}[htb!]
\caption{Performance on UBFC-rPPG when training with synthetic rPPG datasets (SCAMPS \cite{mcduff2022scamps} and UCLA-Synth \cite{wang2022synthetic}). Note that since Contrast-Phys+ can work in a semi-supervised setting, it can be trained on a labeled synthetic dataset and an unlabeled real dataset (unlabeled MMSE-HR\cite{zhang2016multimodal}). The best results are in bold.}
\label{tab:synthetic}
\begin{minipage}[t][][b]{\linewidth}
\centerline{(1) Training with SCAMPS \cite{mcduff2022scamps}}
\begin{tabular}{l|l|ccc}
\hline
\multirow{2}{*}{Methods}        & \multirow{2}{*}{Training sets}                                                    & \multicolumn{3}{c}{Test on UBFC-rPPG}                                                                                      \\ \cline{3-5} 
                                &                                                                                   & \begin{tabular}[c]{@{}c@{}}MAE\\ (bpm)\end{tabular} & \begin{tabular}[c]{@{}c@{}}RMSE\\ (bpm)\end{tabular} & R             \\ \hline
TS-CAN\cite{NEURIPS2020_e1228be4} & labeled SCAMPS                                                                    & 3.62                                                & 6.92                                                 & 0.93          \\ \hline
PhysNet\cite{yu2019remoteBMVC} & labeled SCAMPS                                                                    & 5.40                                                & 10.89                                                & 0.82          \\ \hline
\multirow{5}{*}{Contrast-Phys+} & labeled SCAMPS                                                                    & 0.89                                                & 3.25                                                 & 0.98          \\ \cline{2-5} 
                                & \begin{tabular}[c]{@{}l@{}}labeled SCAMPS\\ and \\ unlabeled MMSE-HR\end{tabular} & \textbf{0.51}                                       & \textbf{2.16}                                        & \textbf{0.99} \\ \cline{2-5} 
& unlabeled MMSE-HR & 1.03 & 2.70 & \textbf{0.99} \\ \hline
\end{tabular}
\end{minipage}
\begin{minipage}[t][][b]{\linewidth}
\vspace{0.1cm}
\centerline{(b) Training with UCLA-Synth \cite{wang2022synthetic}}
\begin{tabular}{l|l|ccc}
\hline
\multirow{2}{*}{Methods} & \multirow{2}{*}{Training sets} & \multicolumn{3}{c}{Test on UBFC-rPPG} \\ \cline{3-5} 
 & & \begin{tabular}[c]{@{}c@{}}MAE\\ (bpm)\end{tabular} & \begin{tabular}[c]{@{}c@{}}RMSE\\ (bpm)\end{tabular} & R \\ \hline
PRN\cite{ba2022style} & labeled UCLA-Synth & 1.09 & 1.99 & 0.83 \\ \hline
PhysNet\cite{yu2019remoteBMVC} & labeled UCLA-Synth & 0.84 & \textbf{1.76} & 0.83 \\ \hline
\multirow{5}{*}{Contrast-Phys+} & labeled UCLA-Synth & 0.74 & 2.29 & \bf 0.99 \\ \cline{2-5} 
 & \begin{tabular}[c]{@{}l@{}}labeled UCLA-Synth\\ and \\ unlabeled MMSE-HR \end{tabular} & \textbf{0.60} & 2.07 & \textbf{0.99} \\ \cline{2-5} 
  & unlabeled MMSE-HR & 1.03 & 2.70 & \textbf{0.99} \\ \hline
\end{tabular}
\end{minipage}
\end{table}

Table \ref{tab:synthetic} shows the HR results on UBFC-rPPG when models are trained with synthetic datasets. We can conclude the following three points corresponding to the three settings. 1) When trained on a labeled synthetic dataset, supervised Contrast-Phys+ outperforms previous supervised methods. 2) Semi-supervised Contrast-Phys+ trained on a labeled synthetic dataset and an unlabeled real dataset (unlabeled MMSE-HR) achieves the best performance while previous supervised methods can only use labeled synthetic data and cannot utilize unlabeled real data for training. 3) Unsupervised Contrast-Phys+ trained on an unlabeled real dataset (unlabeled MMSE-HR) achieves better or comparable performance than previous supervised methods trained on synthetic data. Overall, the results demonstrate that Contrast-Phys+ allows integrating the two solutions via merging both labeled synthetic data and unlabeled real data which can achieve better performance than using either one solution alone.

\subsection{Ablation Study}
\subsubsection{ST-rPPG Block Parameters} \label{sec:ablation_st_rppg_block}

In our ablation study, we investigated the impact of two key parameters of the ST-rPPG block: spatial resolution ($S$) and temporal length ($T$).

Table \ref{tb:sa}(a) presents the heart rate (HR) results for Contrast-Phys+ (0\%) on UBFC-rPPG when varying the spatial resolution (S) of the ST-rPPG block across four levels: 1x1, 2x2, 4x4, and 8x8. It's important to note that 1x1 implies that rPPG spatial similarity is not considered. As evident from the results, the performance with a spatial resolution of 1x1 is inferior to the other resolutions, indicating that rPPG spatial similarity enhances performance. Furthermore, a spatial resolution of 2x2 yields satisfactory results, and larger resolutions do not substantially improve HR estimation. This is because larger resolutions, such as 8x8 or 4x4, provide more rPPG samples, but each block has a smaller receptive field, leading to noisier rPPG samples.

Table \ref{tb:sa}(b) demonstrates the HR results for Contrast-Phys+ (0\%) on UBFC-rPPG while varying the temporal length (T) of the ST-rPPG block across three levels: 5 seconds, 10 seconds, and 30 seconds. The rPPG sample length ($\Delta t$) is the default value (T/2). The results highlight that a temporal length of 10 seconds yields the best performance. A shorter time length (5 seconds) results in coarse PSD estimation, while a longer time length (30 seconds) might violate the conditions for rPPG temporal similarity. As a result, we opted for $S=2$ and $T=10$ seconds in our experiments, as these settings strike a balance and offer optimal performance.

Table \ref{tb:sa}(c) shows the HR results for Contrast-Phys+ (0\%) on UBFC-rPPG while varying the rPPG sample lengths ($\Delta t$) when T = 10s. Three levels of $\Delta t$ are selected: T/4 (2.5s), T/2 (5s), and 3T/4 (7.5s). The findings emphasize that both T/2 and 3T/4 exhibit comparable performance, whereas the shorter T/4 demonstrates lower performance. A shorter $\Delta t$ like T/4 leads to inaccurate PSDs and heart rate peaks, while a longer $\Delta t$ such as T/2 and 3T/4 can offer more precise PSDs, aiding in sample comparisons within contrastive learning. However, long $\Delta t$ like 3T/4 also increases computational costs. Therefore, we adopt T/2 as the default value.

\begin{table}[htb!]
\centering
\caption{Ablation study for ST-rPPG block parameters: (a) HR results of Contrast-Phys+ on UBFC-rPPG with different ST-rPPG block spatial resolutions (S). (b) HR results of Contrast-Phys+ on UBFC-rPPG with different ST-rPPG block time lengths (T) when $\Delta t=T/2$. (c) HR results of Contrast-Phys+ on UBFC-rPPG with different rPPG sample lengths ($\Delta t$) when T = 10s (The best results are in bold.)}
  \begin{minipage}[t][][b]{\linewidth}
    % \centering
    \centerline{(a)}
    % \fontsize{8}{8}\selectfont
    \resizebox{\linewidth}{!}{
    \begin{tabular}{lccc} 
    \toprule
     \begin{tabular}[c]{@{}l@{}} Spatial Resolution (S) \end{tabular}& \begin{tabular}[c]{@{}l@{}} MAE (bpm)\end{tabular} & \begin{tabular}[c]{@{}l@{}} RMSE (bpm)\end{tabular} & R\\
    \midrule
     $1 \times 1$ & 3.14 & 4.06 & 0.963 \\
     $2 \times 2$ & \bf 0.64 & \bf 1.00 & \bf 0.995 \\
     $4 \times 4$ & 0.55 & 1.06 & 0.994 \\
     $8 \times 8$ & 0.60 & 1.09 & 0.993 \\
    \bottomrule
    \end{tabular}
    }
  \end{minipage}
  
  \begin{minipage}[t][][b]{\linewidth}
    % \centering
    \centerline{(b)}
    % \fontsize{8}{9.6}\selectfont
    \resizebox{\linewidth}{!}{
    \begin{tabular}{lccc} 
    \toprule
     \begin{tabular}[c]{@{}l@{}} Time Length (T)\end{tabular}& \begin{tabular}[c]{@{}l@{}} MAE (bpm)\end{tabular} & \begin{tabular}[c]{@{}l@{}} RMSE (bpm)\end{tabular} & R\\
    \midrule
     5s & 0.68 & 1.36 & 0.990 \\
     10s & \bf 0.64 & \bf 1.00 & \bf 0.995 \\
     30s & 1.97 & 3.58 & 0.942 \\
    \bottomrule
    \end{tabular}
    }
  \end{minipage}
  
  \begin{minipage}[t][][b]{\linewidth}
    % \centering
    \centerline{(c)}
    % \fontsize{8}{9.6}\selectfont
    \resizebox{\linewidth}{!}{
    \begin{tabular}{lccc} 
    \toprule
     \begin{tabular}[c]{@{}l@{}} rPPG Sample Length ($\Delta t$)\end{tabular}& \begin{tabular}[c]{@{}l@{}} MAE (bpm)\end{tabular} & \begin{tabular}[c]{@{}l@{}} RMSE (bpm)\end{tabular} & R\\
    \midrule
     T/4 & 0.98  & 1.74  & 0.986  \\
     T/2 & \bf 0.64  & 1.00  & \bf 0.995  \\
     3T/4 & 0.65  & \bf 0.98  &  \bf 0.995 \\
    \bottomrule
    \end{tabular}
    }
  \end{minipage}
  \label{tb:sa}
\end{table}

\subsubsection{rPPG Observations}

\begin{table*}[hbt!]
\centering
\caption{Ablation Study for rPPG Observations on UBFC-rPPG dataset. The best results are in bold.}
  \begin{minipage}[t][][b]{\linewidth}
    \centering
    % \fontsize{8}{8}\selectfont
    \resizebox{\linewidth}{!}{
    \begin{tabular}{ccccccc} 
    \toprule
     \begin{tabular}[c]{@{}l@{}} rPPG Spatial\\Similarity \end{tabular}& \begin{tabular}[c]{@{}l@{}} rPPG Temporal\\Similarity \end{tabular} & \begin{tabular}[c]{@{}l@{}} rPPG Cross-video\\Dissimilarity \end{tabular} & \begin{tabular}[c]{@{}l@{}} HR Range Constraint \end{tabular} & \begin{tabular}[c]{@{}l@{}} MAE (bpm) \end{tabular} & \begin{tabular}[c]{@{}l@{}} RMSE (bpm) \end{tabular} & \begin{tabular}[c]{@{}l@{}} R \end{tabular}\\
    \midrule
      \checkmark &\checkmark  &\checkmark  &  & 39.66 & 44.49 & -0.401 \\
      \checkmark & \checkmark & &\checkmark & 22.11 & 33.84 & 0.281   \\
      \checkmark&  &\checkmark  &\checkmark &1.26 & 3.64 & 0.948    \\
      &\checkmark  &\checkmark  &\checkmark & 3.14 & 4.06 & 0.963 \\
      \checkmark &\checkmark  &\checkmark  &\checkmark  & \bf 0.64 & \bf 1.00 & \bf 0.995 \\
    \bottomrule
    \end{tabular}
    }
  \end{minipage}
  \label{tb:ablation_rppg}
\end{table*}

In our ablation study, we examined the individual impact of each of the four rPPG observations on the performance of Contrast-Phys+. These observations include rPPG spatial and temporal similarity (represented by rPPG spatial and temporal sampling), rPPG cross-video dissimilarity (represented by the rPPG-rPPG negative loss $L_n^{RR}$), and the HR range constraint (utilizing PSDs in the HR frequency range).

Table \ref{tb:ablation_rppg} showcases the results for Contrast-Phys+ (0\%) when one of the rPPG observations is removed, as well as the results when all observations are utilized. The findings indicate that Contrast-Phys+ achieves its best performance when all rPPG observations are enabled. When rPPG spatial or temporal similarity is disabled, the performance experiences a slight decrease. However, when rPPG cross-video dissimilarity or the HR range constraint is disabled, the performance deteriorates significantly. The HR range constraint plays a crucial role in preventing the model from learning irrelevant periodic noises, such as light flickering, which can interfere with accurate HR estimation. Additionally, rPPG cross-video dissimilarity, represented by the rPPG-rPPG negative loss $L_n^{RR}$, is essential in contrastive learning as it prevents the model from collapsing into trivial solutions, as discussed in \cite{hadsell2006dimensionality}.

These results underscore the importance of all four rPPG observations in enhancing the performance of Contrast-Phys+ and emphasize their individual contributions to accurate and robust rPPG signal extraction.

\subsubsection{The Influence of GT Signals}\label{sec:GT_influence}

\textbf{GT-related Losses.} We conducted an ablation study to assess the influence of GT-related losses on our model's performance. Table \ref{tb:ablation_gt} presents the results of the ablation study performed on the MMSE-HR dataset using Contrast-Phys+ (100\%). When we exclude all GT-related terms, the model effectively undergoes unsupervised training, resulting in the lowest performance. However, when we include only the GT-rPPG negative term, the model's performance improves, as it generates more negative pairs from both GT signals and ST-rPPG blocks. Subsequently, utilizing solely the GT-rPPG positive term further enhances performance, as it enforces consistency between ST-rPPG blocks and their corresponding GT signals, effectively incorporating GT information into the model's training. The combined use of both terms yields the highest performance, which is the top-performing configuration.

\begin{table}[hbt!]
\centering
\caption{Ablation Study for GT-related Positive Loss Term $L_p^{GR}$ and Negative Loss Term $L_n^{GR}$ on MMSE-HR dataset. The best results are in bold.}
  \begin{minipage}[t][][b]{\linewidth}
    % \centering
    % \fontsize{8}{8}\selectfont
    \resizebox{\linewidth}{!}{
    \begin{tabular}{ccccc} 
    \toprule
     \begin{tabular}[c]{@{}l@{}} $L_p^{GR}$ \end{tabular}& \begin{tabular}[c]{@{}l@{}} $L_n^{GR}$ \end{tabular} & \begin{tabular}[c]{@{}l@{}} MAE (bpm) \end{tabular} & \begin{tabular}[c]{@{}l@{}} RMSE (bpm) \end{tabular} & \begin{tabular}[c]{@{}l@{}} R \end{tabular}\\
    \midrule
        &  & 1.82 & 6.69 & 0.87 \\
        &\checkmark & 1.77 & 5.30 & 0.88   \\
      \checkmark& &1.39 & 4.46 & 0.91    \\
      \checkmark  &\checkmark & \bf 1.11 & \bf 3.83 & \bf 0.96 \\
    \bottomrule
    \end{tabular}
    }
  \end{minipage}
  \label{tb:ablation_gt}
\end{table}

\begin{figure}[hbt!]
\centering
\begin{minipage}[b]{0.49\linewidth}
  \centering
  \centerline{\includegraphics[width=\linewidth]{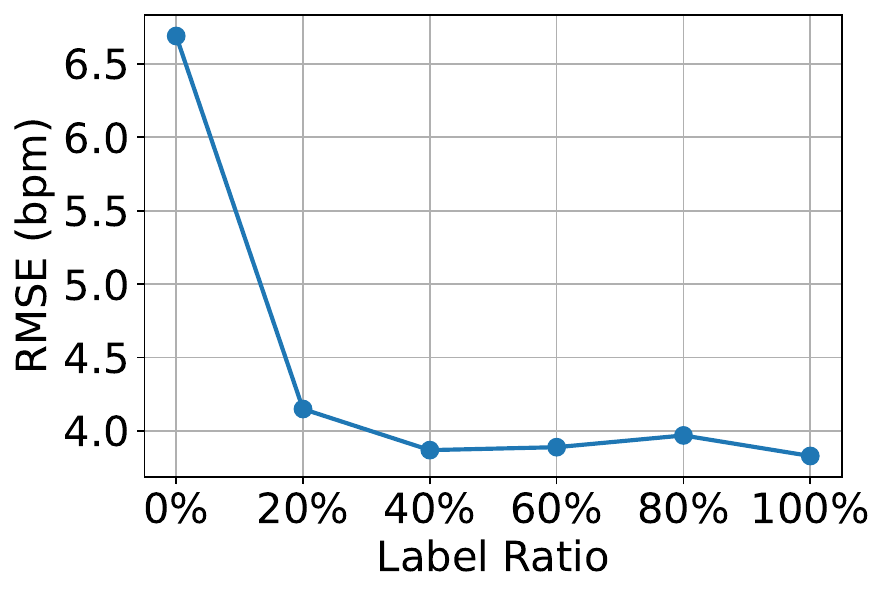}}
  \centerline{(a) RMSE}
\end{minipage}
\begin{minipage}[b]{0.49\linewidth}
  \centering
  \centerline{\includegraphics[width=\linewidth]{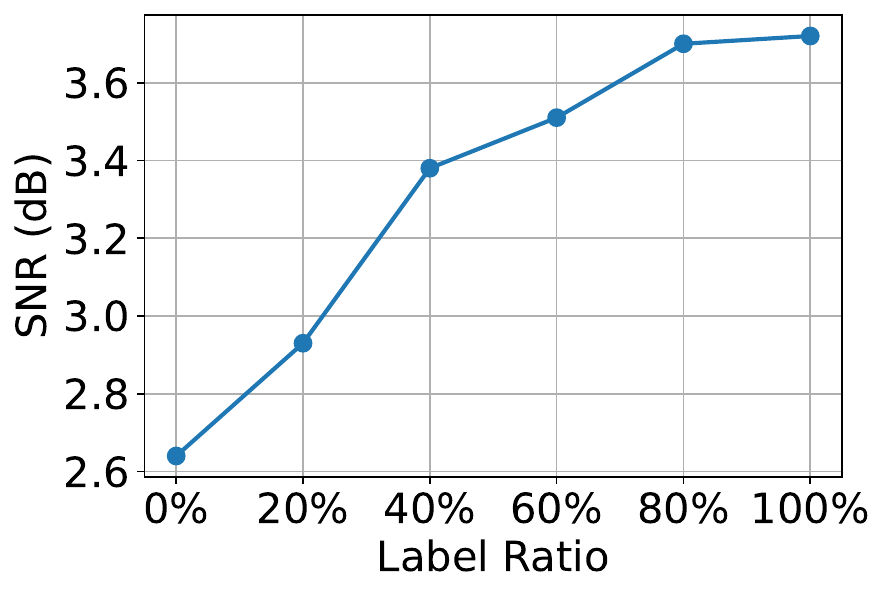}}
  \centerline{(b) SNR}
\end{minipage}
\caption{rPPG measurement performance ((a) RMSE and (b) SNR) with respect to label ratios.}
\label{fig:label_ratio}
\end{figure}

\textbf{GT Signal Ratios.} Since Contrast-Phys+ is capable of adapting to different availability of data labels, we conducted an ablation study to examine the impact of different GT signal ratios. Specifically, we trained Contrast-Phys+ using 0\%, 20\%, 40\%, 60\%, 80\%, and 100\% labels from the MMSE-HR dataset. The performance variation of Contrast-Phys+ under different label ratios is illustrated in Fig. \ref{fig:label_ratio}. Regarding RMSE, the performance reaches a plateau at 40\% label ratio, and the HR error does not significantly decrease when using more than 40\% labels. On the other hand, SNR, which serves as a metric for rPPG signal quality, exhibits continuous improvement with an increasing number of labels. These findings suggest that while employing more labels (beyond 40\%) may not lead to a substantial reduction in HR measurement error, they do contribute to refining the quality of the output rPPG signals. We will further demonstrate this through waveform visualization in Sec. \ref{sec:rppg_waveform}.

\subsection{Statistical Validation for rPPG Observations}

\begin{figure*}[hbt!]
\centering
\begin{minipage}[b]{0.19\linewidth}
  \centering
  \centerline{\includegraphics[width=\linewidth]{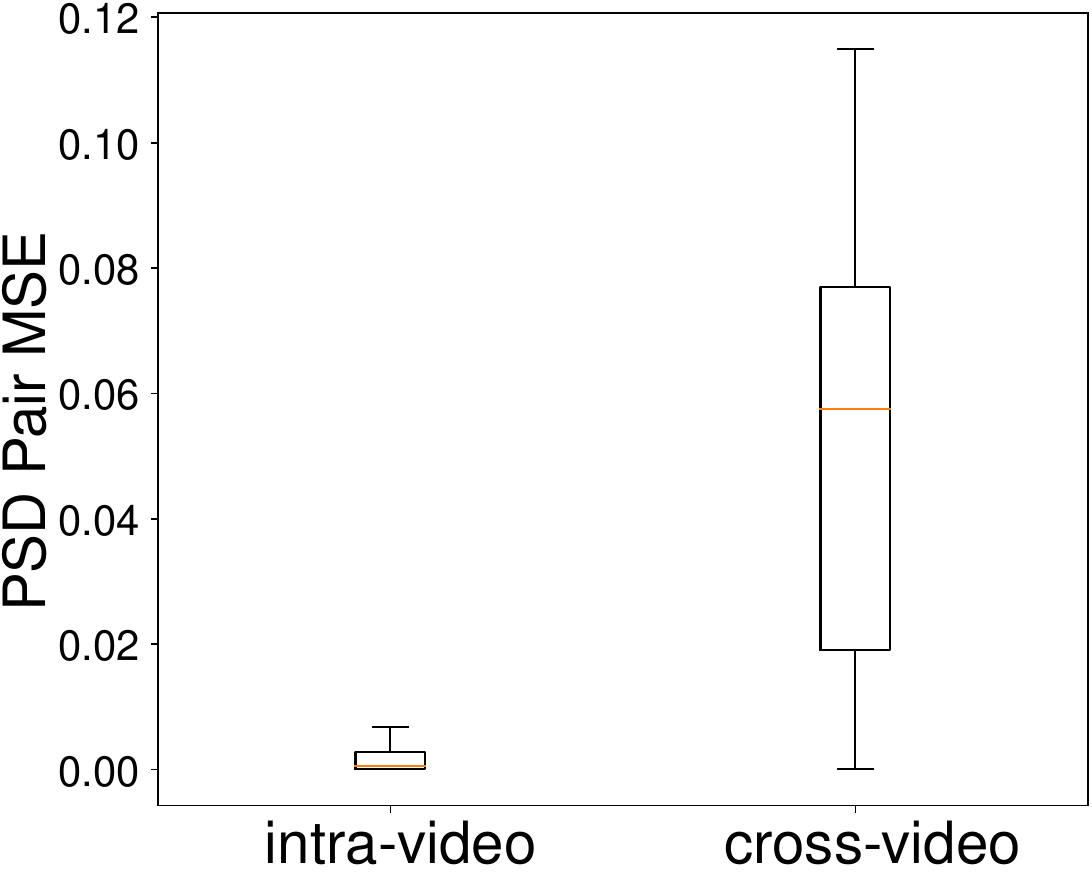}}
  \centerline{\footnotesize (a) PURE}
\end{minipage}
\begin{minipage}[b]{0.19\linewidth}
  \centering
  \centerline{\includegraphics[width=\linewidth]{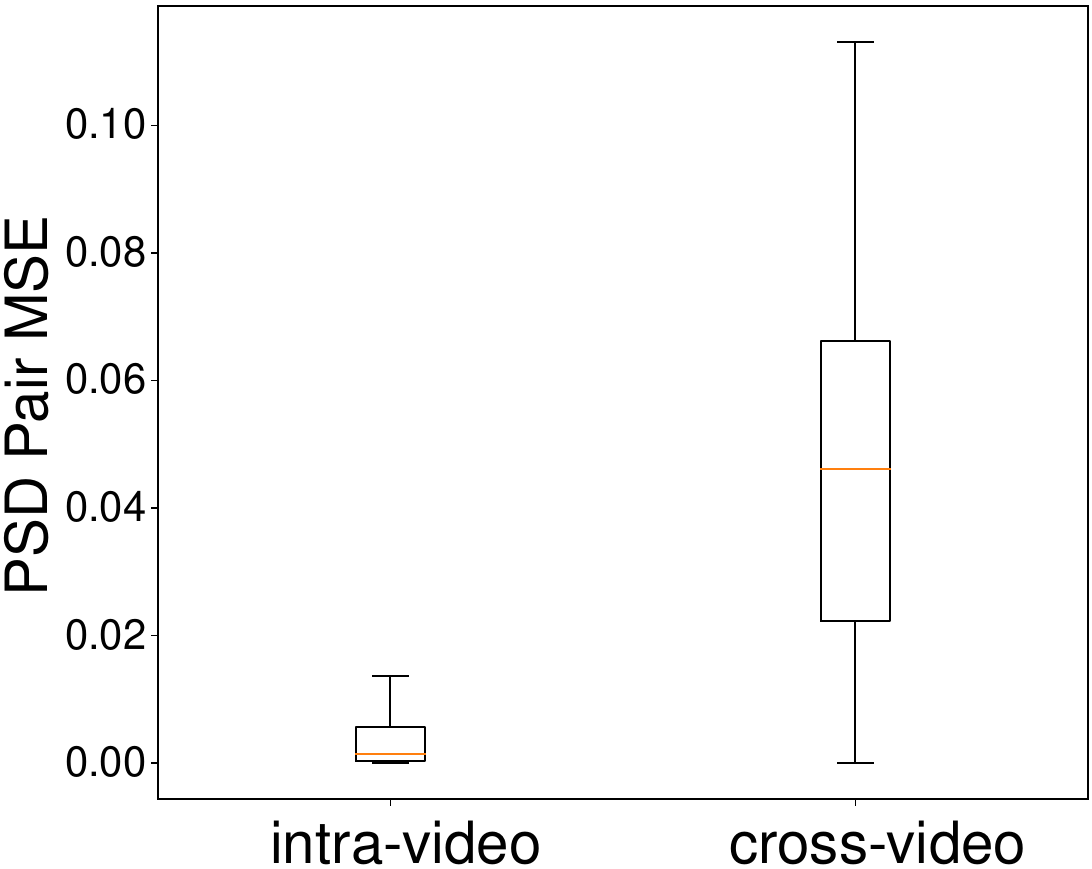}}
  \centerline{\footnotesize (b) UBFC}
\end{minipage}
\begin{minipage}[b]{0.19\linewidth}
  \centering
  \centerline{\includegraphics[width=\linewidth]{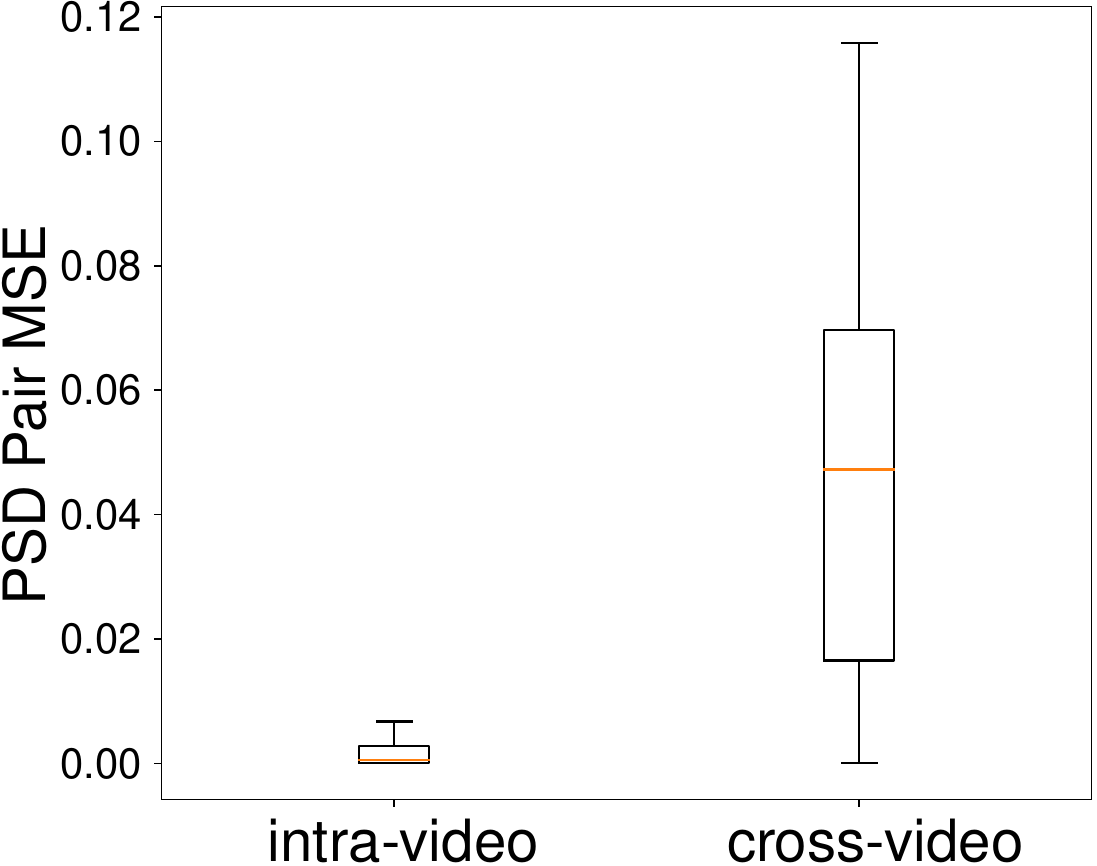}}
  \centerline{\footnotesize (c) OBF}
\end{minipage}
\begin{minipage}[b]{0.19\linewidth}
  \centering
  \centerline{\includegraphics[width=\linewidth]{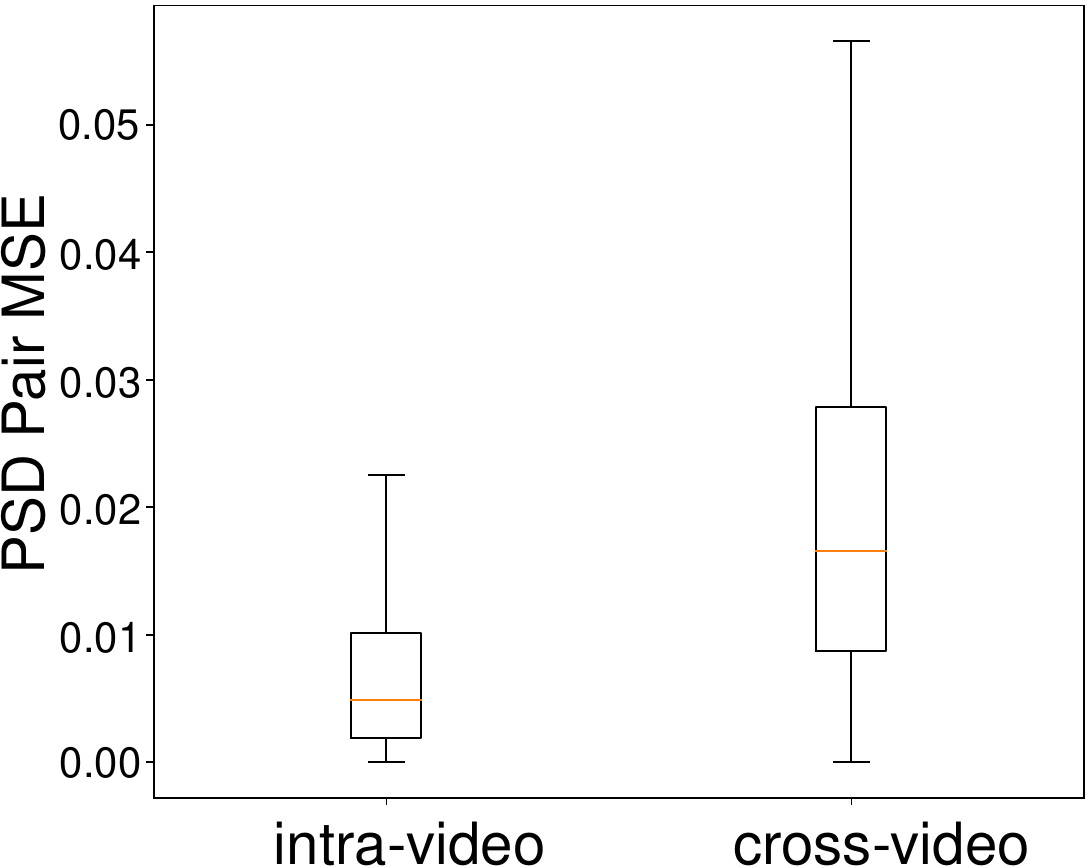}}
  \centerline{\footnotesize (d) MR-NIRP}
\end{minipage}
\begin{minipage}[b]{0.19\linewidth}
  \centering
  \centerline{\includegraphics[width=\linewidth]{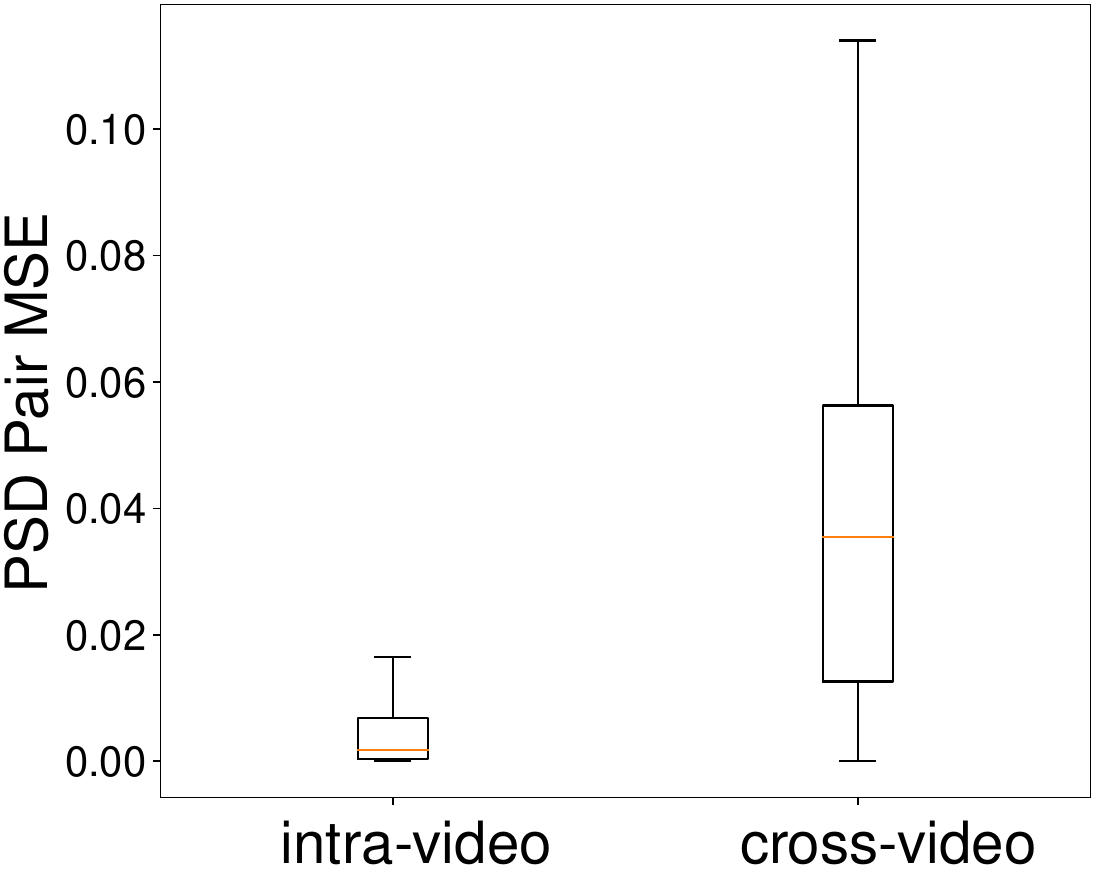}}
  \centerline{\footnotesize (e) MMSE-HR}
\end{minipage}
\caption{Boxplots of PSD pair MSE for intra-video and cross-video for rPPG datasets: (a) PURE, (b) UBFC, (c) OBF, (d) MR-NIRP, and (e) MMSE-HR.}
\label{fig:psd_intra_cross}
\end{figure*}

We conducted a statistical analysis to validate both the spatiotemporal similarity of rPPG signals within the same video (referred to as "intra-video") and the dissimilarity of rPPG signals between different videos (referred to as "cross-video"). The spatiotemporal similarity of rPPG signals refers to the similarity in the PSDs of rPPG signals measured at different spatiotemporal locations within the same video. Conversely, the cross-video rPPG dissimilarity refers to the differences in the PSDs of rPPG signals measured at different spatiotemporal locations between two different videos.

To quantify these observations, we calculated the mean squared errors (MSE) of PSD pairs for both the intra-video and cross-video cases. Figure \ref{fig:psd_intra_cross} illustrates that the PSD pair MSE for the intra-video case is significantly smaller compared to the PSD pair MSE for the cross-video case. To assess the significance of these differences, we employed the two-sample Kolmogorov-Smirnov test \cite{hodges1958significance,mcduff2014remote}. The results indicate that the PSD pair MSE for the cross-video case is significantly higher than for the intra-video case ($p<0.001$) across all five rPPG datasets. These statistical test results provide solid evidence supporting the validity of both the rPPG spatiotemporal similarity and the cross-video rPPG dissimilarity observations.

\begin{figure*}[hbt!]
\centering
\begin{minipage}[b]{\linewidth}
  \centering
  \centerline{\includegraphics[width=\linewidth]{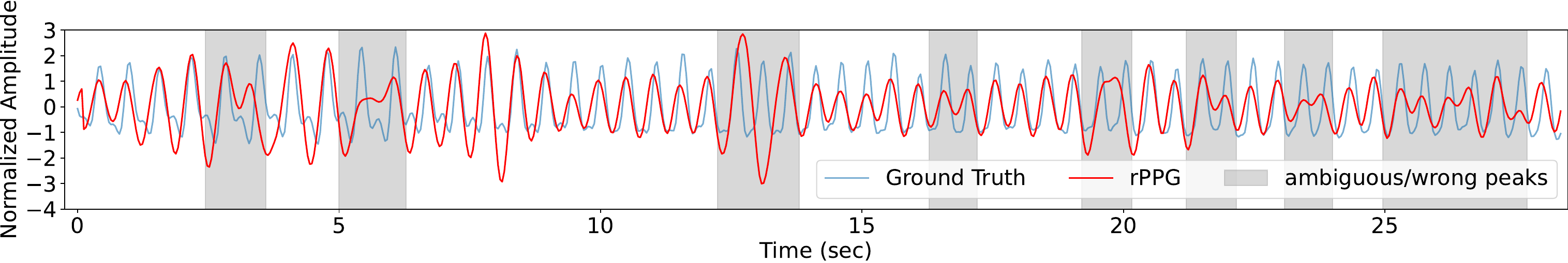}}
  \centerline{(a) 0\%}
\end{minipage}

\begin{minipage}[b]{\linewidth}
  \centering
  \centerline{\includegraphics[width=\linewidth]{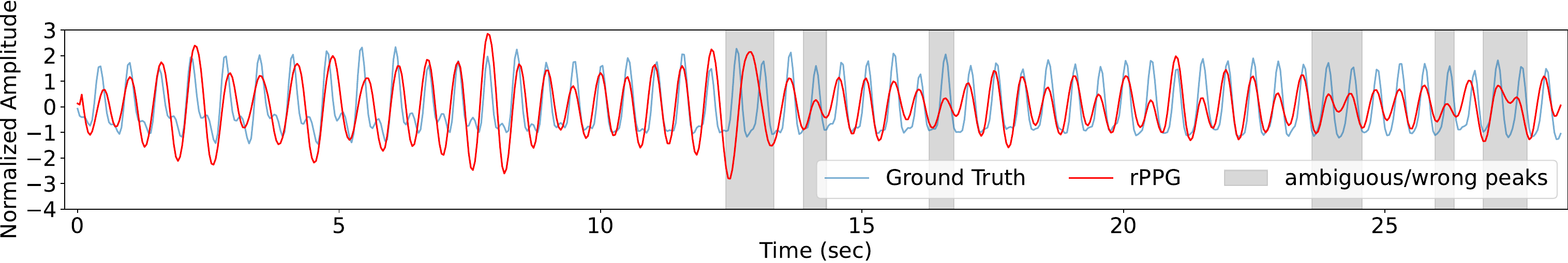}}
  \centerline{(b) 20\%}
\end{minipage}

\begin{minipage}[b]{\linewidth}
  \centering
  \centerline{\includegraphics[width=\linewidth]{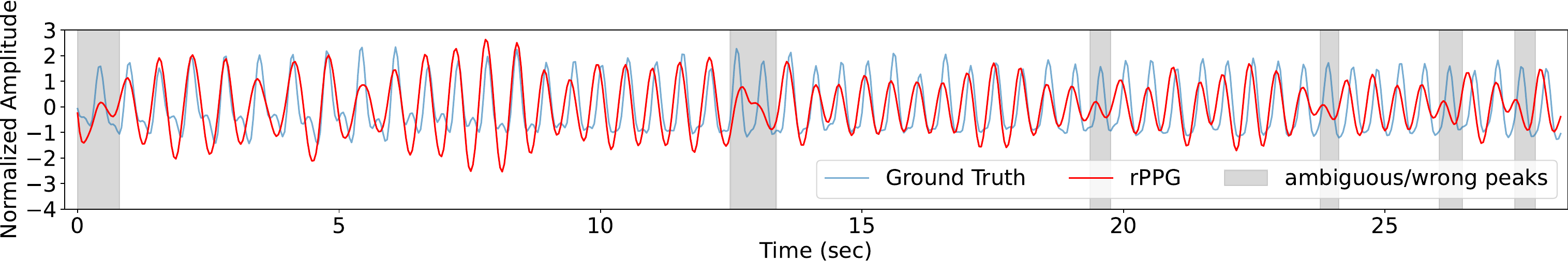}}
  \centerline{(c) 60\%}
\end{minipage}

\begin{minipage}[b]{\linewidth}
  \centering
  \centerline{\includegraphics[width=\linewidth]{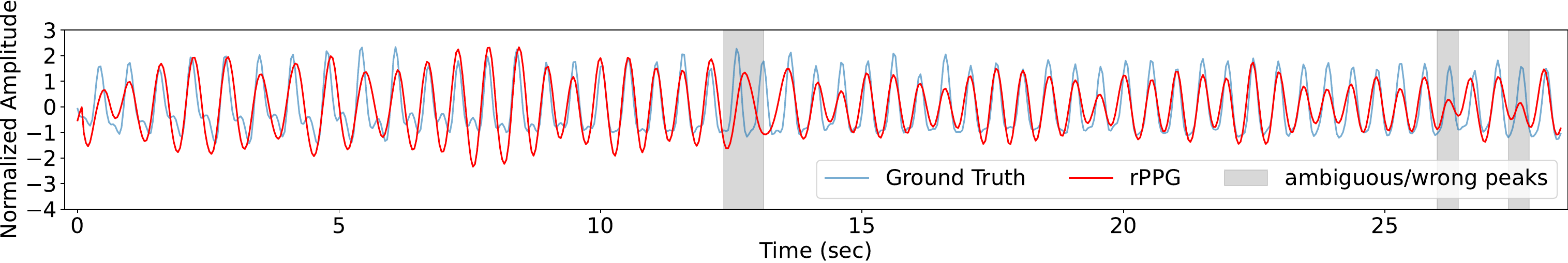}}
  \centerline{(d) 100\%}
\end{minipage}
\caption{rPPG waveforms from Contrast-Phys+ trained with different label ratios: (a) 0\%, (b) 20\%, (c) 60\%, and (d) 100\%. The ambiguous/wrong peaks from rPPG are highlighted in gray areas.}
\label{fig:rppg_waveform_label_ratio}
\end{figure*}

\subsection{Running Speed}

We conducted experiments to compare the running speed of Contrast-Phys+ and Gideon2021 \cite{gideon2021way}. During training, the running speed of Contrast-Phys+ (0\%) was measured at \textbf{802.45} frames per second (fps), while Gideon2021 achieved a speed of \textbf{387.87} fps, which is approximately half of Contrast-Phys+'s speed. This significant difference in speed can be attributed to the different method designs employed by the two models. In Gideon2021, the input video is fed into the model twice, first as the original video and then as a temporally resampled video, resulting in double computation. On the other hand, Contrast-Phys+ only requires the input video to be fed into the model once, leading to a substantial decrease in computational cost. Additionally, the running speed of Contrast-Phys+ for label ratios of 60\% and 100\% was measured at \textbf{792.70} fps and \textbf{776.19} fps, respectively. When compared to Contrast-Phys+ (0\%) (802.45 fps), incorporating GT signals in Contrast-Phys+ (60\%, 100\%) only resulted in a slight decrease in speed.

Furthermore, we compared the convergence speed using the metric of Irrelevant power ratio (IPR). IPR is used in \cite{gideon2021way} to evaluate signal quality during training with lower values indicating higher signal quality. More details about IPR can be found in the supplementary materials. Figure \ref{fig:ipr} illustrates the IPR values over time during training on the OBF dataset. The results demonstrate that Contrast-Phys+ achieves faster convergence to a lower IPR compared to Gideon2021. While Contrast-Phys+ (60\%, 100\%) takes slightly longer to reach the lowest IPR compared to Contrast-Phys+ (0\%), it ultimately achieves a lower IPR due to its ability to utilize GT signals to further enhance the rPPG signal quality.

\begin{figure}[hbt!]
\centering
\begin{minipage}[b]{0.8\linewidth}
  \centering
  \centerline{\includegraphics[width=\linewidth]{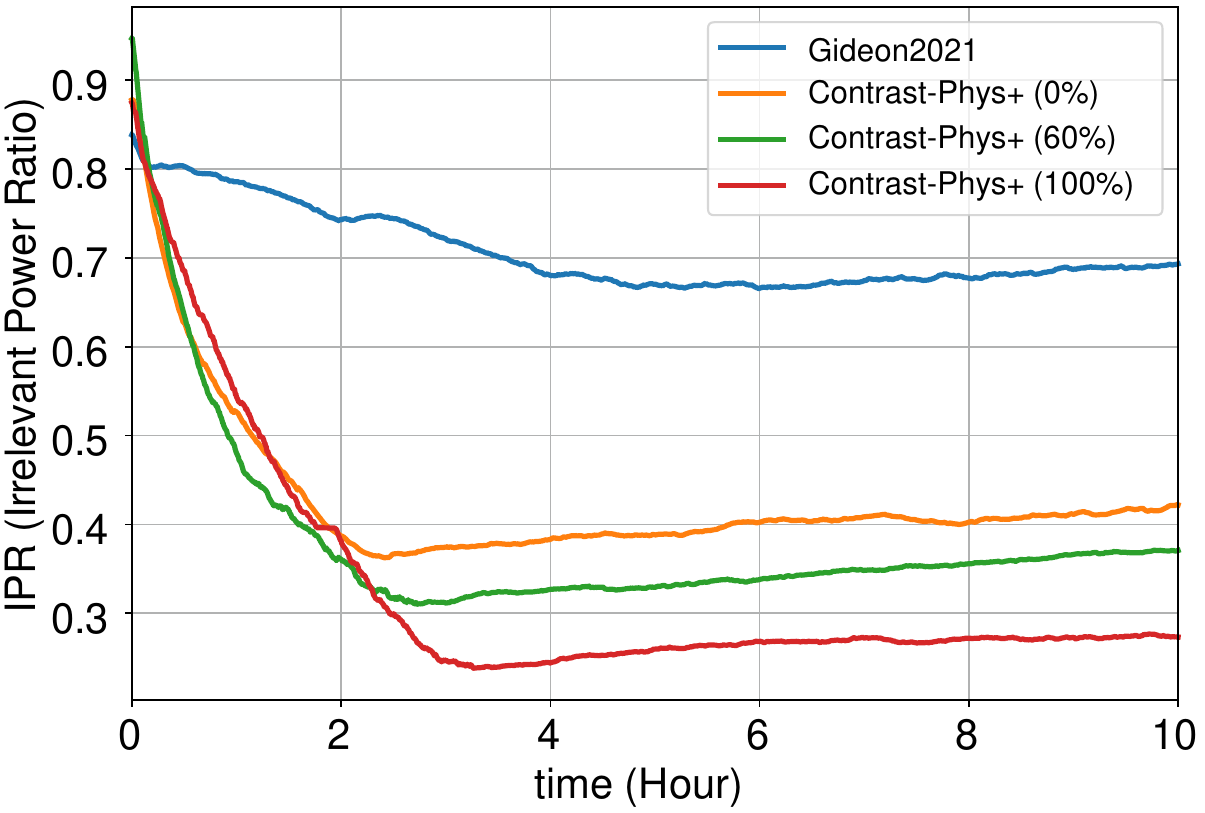}}
\end{minipage}
\caption{Irrelevant power ratio (IPR) with respect to training time.}
\label{fig:ipr}
\end{figure}

\subsection{Result Visualization}

\subsubsection{Saliency Maps}

\begin{figure}[hbt!]
\centering
\begin{minipage}[b]{\linewidth}
  \centering
  \centerline{\includegraphics[width=\linewidth]{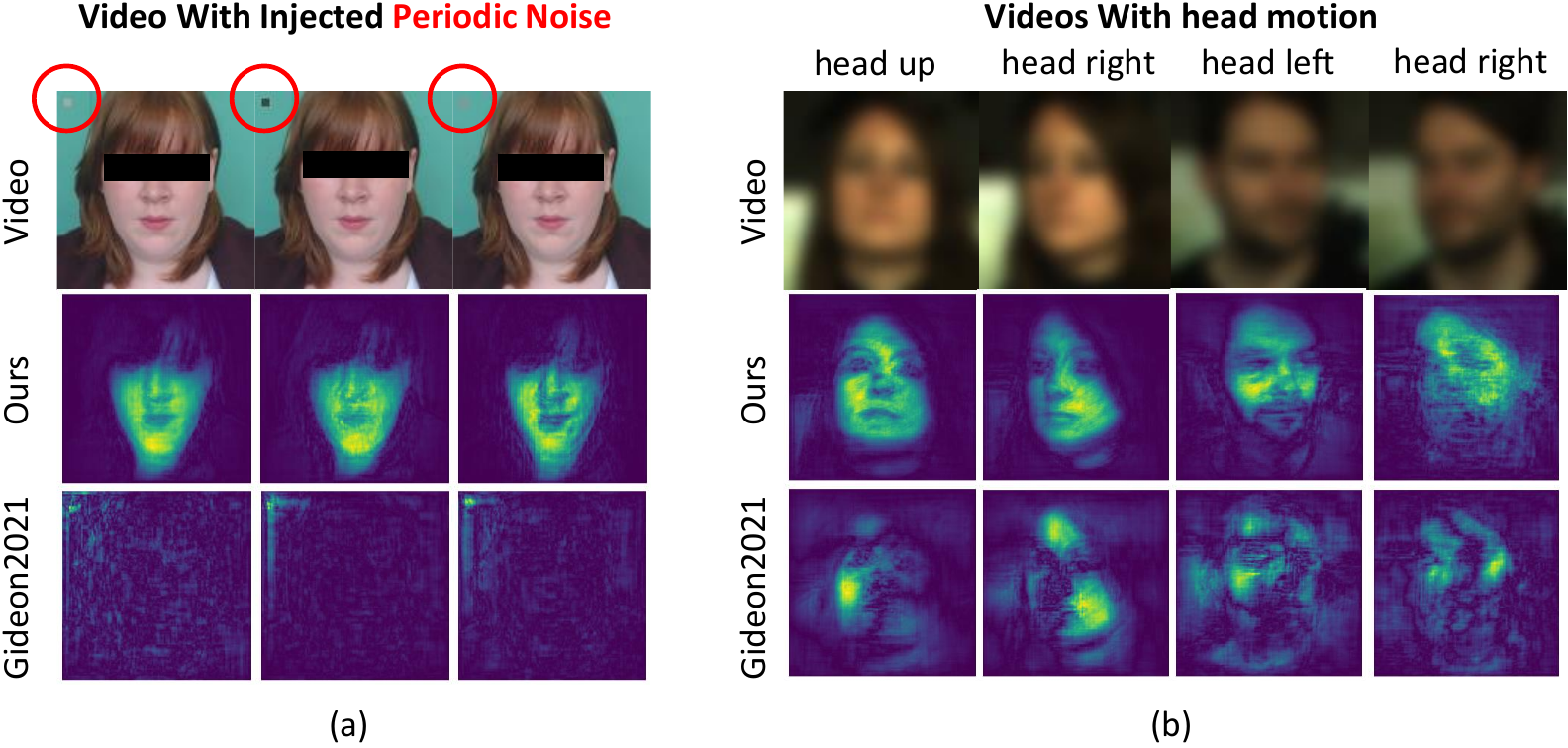}}
\end{minipage}
\caption{Saliency maps for Contrast-Phys+ (0\%) and Gideon2021 \cite{gideon2021way} (a) In this scenario, we introduce a random flashing block with a heart rate (HR) range between 40-250 bpm in the top-left corner of all UBFC-rPPG videos. Both Contrast-Phys+ (0\%) and Gideon2021 are trained on these videos with the added noise. The saliency maps of Contrast-Phys+ (0\%) exhibit a strong response in facial regions, indicating its focus on these areas. On the other hand, Gideon2021 primarily focuses on the injected periodic noise. The quantitative results in Table \ref{tb:noise} further support the robustness of Contrast-Phys+ (0\%) to the noise. (b) For this scenario, we select moments of head motion from the PURE dataset (Video frames shown here are blurred due to privacy concerns.) and generate the corresponding saliency maps for Contrast-Phys+ (0\%) and Gideon2021.}
\label{fig:saliency}
\end{figure}

To demonstrate the interpretability of Contrast-Phys+, we present saliency maps. These saliency maps are generated using a gradient-based method proposed in \cite{simonyan2013deep}. We keep the weights of the trained model fixed and calculate the gradient of the Pearson correlation with respect to the input video. More detailed information can be found in the supplementary materials. Saliency maps are useful for highlighting the spatial regions that contribute to the estimation of rPPG signals by the model. A saliency map of a good rPPG model should exhibit a strong response in skin regions, as demonstrated in previous works such as \cite{yu2019remoteBMVC,yu2019remote,gideon2021way,chen2018deepphys,nowara2021benefit}.

\begin{table}[hbt!]
    \centering
    \caption{HR results trained on UBFC-rPPG with/without injected periodic noise shown in Fig. \ref{fig:saliency}(a).}
    % \scriptsize
    % \fontsize{8}{8}\selectfont
    % \setlength{\tabcolsep}{1.5pt}
    \resizebox{\linewidth}{!}{
    \begin{tabular}{lcccc} 
    \toprule
    \begin{tabular}[c]{@{}l@{}} Methods\end{tabular} & \begin{tabular}[c]{@{}l@{}} Injected \\ Periodic \\ Noise\end{tabular}& \begin{tabular}[c]{@{}l@{}} MAE\\ (bpm)\end{tabular} & \begin{tabular}[c]{@{}l@{}} RMSE\\ (bpm)\end{tabular} & R\\
    \midrule
    \multirow{2}{*}{\begin{tabular}[c]{@{}l@{}}Gideon2021 \cite{gideon2021way}\end{tabular}}
    & w/o & 1.85 & 4.28 & 0.939 \\
    & w/ & 22.47 & 25.41 & 0.244 \\
    \midrule
    \multirow{2}{*}{\begin{tabular}[c]{@{}l@{}}\bf Contrast-Phys+ (0\%)\end{tabular}}
    & w/o & 0.64 & 1.00 & 0.995 \\
    & w/ & 0.74 & 1.34 & 0.991 \\
    \bottomrule
    \end{tabular}
    }
    \label{tb:noise}
\end{table}

Fig. \ref{fig:saliency} presents saliency maps in two scenarios to showcase the robustness of our method against interferences: 1) when periodic noise is manually injected, and 2) when head motion is involved. In the presence of a periodic noise patch injected into the upper-left corner of the videos, Contrast-Phys+ remains unaffected by the noise and continues to focus on skin areas. In contrast, Gideon2021 is completely distracted by the noise block. We also evaluate the performance of both methods on UBFC-rPPG videos with the injected noise, and the results are summarized in Table \ref{tb:noise}. These results align with the saliency map analysis, confirming that Contrast-Phys+ is not impacted by the periodic noise, while Gideon2021 fails to handle it effectively. The robustness of Contrast-Phys+ to noise can be attributed to the rPPG spatial similarity constraint, which helps to filter out noise. Fig. \ref{fig:saliency}(b) displays the saliency maps when head motion is involved. The saliency maps of Contrast-Phys+ primarily focus on and activate skin areas, indicating its ability to handle head motions effectively. In contrast, the saliency maps of Gideon2021 exhibit noise and are scattered, covering only partial facial areas during head motions.

\subsubsection{rPPG Waveforms} \label{sec:rppg_waveform}

Fig. \ref{fig:rppg_waveform_label_ratio} displays the rPPG waveforms obtained from Contrast-Phys+ trained with different label ratios on the MMSE-HR dataset. As more labels are available during training, ranging from 0\% to 100\%, the rPPG waveforms become more similar to the ground truth (GT) signal and exhibit fewer ambiguous or incorrect peaks. The waveform corresponding to the 0\% label ratio contains noisy components highlighted by gray areas indicating ambiguous or incorrect peaks. In contrast, the waveform at the 100\% label ratio is well aligned with the GT signal, with almost all peaks clearly distinguishable. The presence of distinguishable peaks in the rPPG waveform also facilitates the accurate calculation of HRV. The visualization of rPPG waveforms demonstrates that incorporating more labels during the training of Contrast-Phys+ improves the quality of the rPPG signal. This finding is consistent with the signal-to-noise ratio (SNR) results discussed in Sec. \ref{sec:GT_influence}.

\section{Conclusion}\label{sec13}

We propose Contrast-Phys+, which can be trained in unsupervised and weakly-supervised settings and achieve accurate rPPG measurement. Contrast-Phys+ is based on four rPPG observations and utilizes spatiotemporal contrast to enable unsupervised and weakly-supervised learning including missing and unsynchronized GT signals, or even no labels. By combining rPPG prior knowledge and additional GT information, Contrast-Phys+ outperforms both unsupervised and supervised state-of-the-art methods and achieves good generalization to unseen data. Besides, the proposed method is robust against noise interference and computationally efficient. For future studies, the proposed method can be extended to learn other periodic signals such as respiration signals.

\ifCLASSOPTIONcompsoc
  \section*{Acknowledgments}
\else
  \section*{Acknowledgment}
\fi
This work was supported by the Research Council of Finland (former Academy of Finland) Academy Professor project EmotionAI (grants 336116, 345122), ICT 2023 project TrustFace (grant 345948), the University of Oulu \& Research Council of Finland Profi 7 (grant 352788), and by Infotech Oulu. The work was also supported by the Spearhead project 'Gaze on Lips' funded by the Eudaimonia Institute of the University of Oulu, Finland. The authors also acknowledge CSC-IT Center for Science, Finland, for providing computational resources.

\ifCLASSOPTIONcaptionsoff
  \newpage
\fi



\bibliographystyle{IEEEtran}
\bibliography{IEEEabrv,ref}



% \end{thebibliography}

% \begin{IEEEbiography}[{\includegraphics[width=1in,height=1.25in,clip,keepaspectratio]{mshell}}]{Michael Shell}

\begin{IEEEbiography}[{\includegraphics[width=1in,height=1.25in,clip,keepaspectratio]{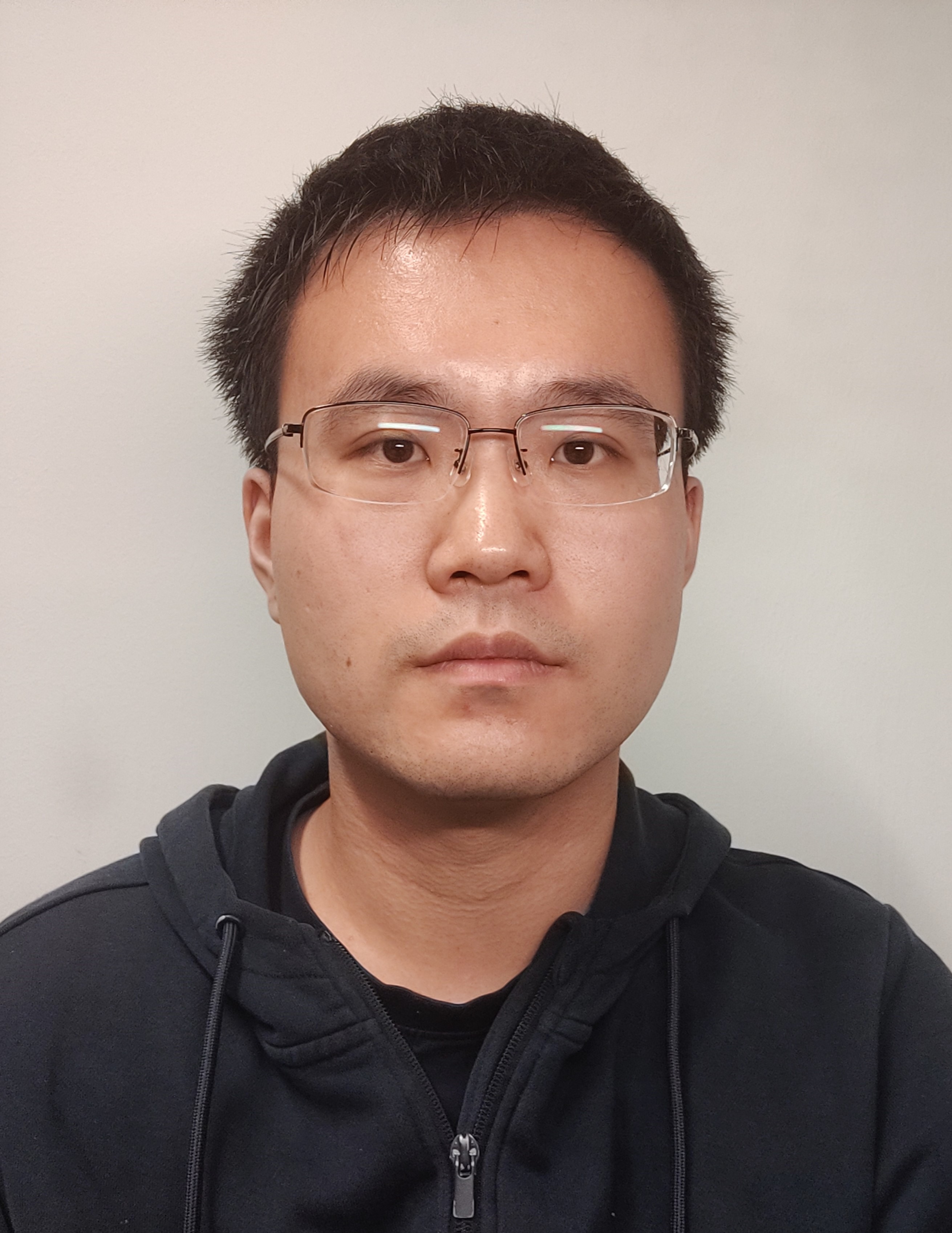}}]{Zhaodong Sun}
is a final-year doctoral student at the Center for Machine Vision and Signal Analysis, University of Oulu. He received M.Sc. and B.E from Swiss Federal Institute of Technology Lausanne and University of Electronic Science and Technology of China in 2020 and 2018. His research interests include computer vision, biomedical signal processing, remote physiological measurement, and affective computing.
\end{IEEEbiography}

\begin{IEEEbiography}[{\includegraphics[width=1in,height=1.25in,clip,keepaspectratio]{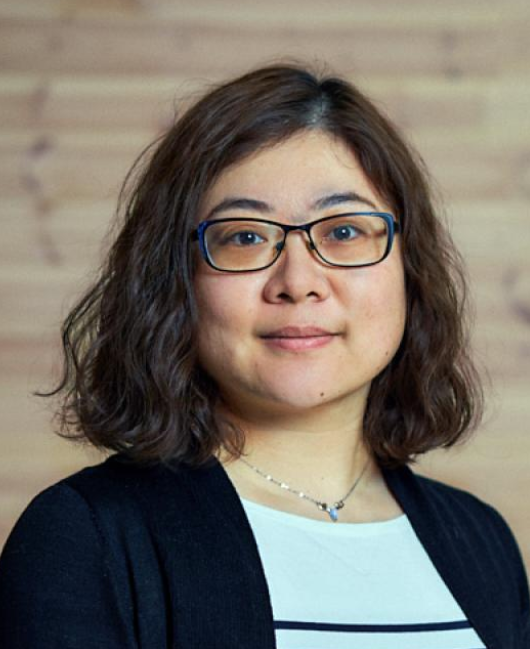}}]{Xiaobai Li}
(Senior Member, IEEE) received the BSc degree in psychology from Peking University, the MSc degree in biophysics from the Chinese Academy of Science, and the PhD degree in computer science from the University of Oulu. She is currently a ZJU100 professor with the School of Cyber Science and Technology, Zhejiang University, and she is also an Adjunct Professor with the Center for Machine Vision and Signal Analysis, University of Oulu. Her research of interests include facial expression recognition, micro-expression analysis, remote physiological signal measurement from facial videos, and related applications in affective computing, healthcare and biometrics. She is an associate editor of IEEE Transactions on Circuits and Systems for Video Technology, Frontiers in Psychology, and Image and Vision Computing. She was a co-chair of several international workshops in CVPR, ICCV, FG, and ACM Multimedia.

\end{IEEEbiography}


%\newpage



%\vfill


\end{document}